\documentclass[10pt,journal,cspaper,compsoc]{IEEEtran}
\usepackage[cmex10]{amsmath}
\usepackage{subfigure,cite,bbm,amssymb,multirow,amsthm,bm}
\usepackage{mathrsfs}
\usepackage{graphicx}
\usepackage{epstopdf}
\usepackage{algorithm}
\usepackage{algorithmic}
\usepackage{url}
\usepackage{array}
\usepackage{mdwmath}
\usepackage{mdwtab}
\usepackage{cite}
\usepackage[font={footnotesize}]{caption}
\usepackage{eqparbox}
\def \H {{\mathcal{H}}}

\def \X {{\mathbf{X}}}
\def \A {{\mathbf{A}}}
\def \D {{\mathbf{D}}}
\def \L {{\mathbf{L}}}

\def \x {{\mathbf{x}}}
\def \y {{\mathbf{y}}}
\def \Y {{\mathbf{Y}}}

\def \w {{\mathbf{w}}}

\def \e {{\mathbf{e}}}

\def \h {{\mathbf{h}}}
\def \H {{\mathbf{H}}}
\def \ones {{\mathbf{1}}}

\newtheorem{proposition}{Proposition}
\newtheorem{remark}{Remark}

\DeclareMathAlphabet{\mathpzc}{OT1}{pzc}{m}{it}

\begin{document}
\title{Embedding Graphs under Centrality Constraints for Network Visualization}
\author{Brian~Baingana,~\IEEEmembership{Student Member,~IEEE,}
        and~Georgios B.~Giannakis,~\IEEEmembership{Fellow,~IEEE}
\IEEEcompsocitemizethanks{
\IEEEcompsocthanksitem Partial results were presented at the $38$th
International Conf. on Acoust., Speech and Signal Proc., 
Vancouver, Canada, May 2013 (see also \cite{baingana}).
\IEEEcompsocthanksitem B. Baingana and G. B. Giannakis are
with the Department of Electrical and Computer Engineering, University of
Minnesota, Minneapolis,MN, 55455.E-mail: \{baing011, georgios\}@umn.edu.}
\thanks{}}

%

\IEEEcompsoctitleabstractindextext{%
\begin{abstract}
\indent Visual rendering of graphs is a key task in the mapping of
complex network data. Although most graph drawing algorithms
emphasize aesthetic appeal, certain applications such as travel-time
maps place more importance on visualization of structural network
properties. The present paper advocates two graph embedding approaches
with centrality considerations to comply with node hierarchy. The
problem is formulated first as one of constrained multi-dimensional
scaling (MDS), and it is solved via block coordinate descent
iterations with successive approximations and guaranteed convergence
to a KKT point. In addition, a regularization term enforcing graph
smoothness is incorporated with the goal of reducing edge crossings.
A second approach leverages the locally-linear embedding (LLE) 
algorithm which assumes that the graph encodes data sampled from
a low-dimensional manifold. Closed-form solutions to the resulting
centrality-constrained optimization problems are determined yielding
meaningful embeddings. Experimental results demonstrate the efficacy
of both approaches, especially for visualizing large networks
 on the order of thousands of nodes.
\end{abstract}

}

\maketitle

\IEEEdisplaynotcompsoctitleabstractindextext

\IEEEpeerreviewmaketitle

\section{Introduction}
\label{introduction}

\IEEEPARstart{G}{raphs} offer a valuable means of encoding
relational information between entities of complex systems, arising
in modern communications, transportation and social networks, among
others~\cite{netsci},~\cite{newman}. Although graph embedding
serves a number of tasks including data compression and
a means to geometrically solve computationally hard tasks,
information visualization 
remains its most popular application. 
However, the rising complexity and volume of networked 
data presents new opportunities 
and challenges for visual analysis tools 
that capture global patterns 
and use visual metaphors to convey meaningful 
structural information like hierarchy, similarity and
natural communities \cite{harrison}. 

Most traditional visualization algorithms trade-off
the clarity of structural characteristics of the underlying data
for aesthetic requirements like minimal edge crossing
and fixed inter-node distance; e.g.,~\cite{cise, ask, ignatowicz, 
papakostas, reingold, supowit,
wetherell, biedl, tunkelang}. Although efficient for relatively small
graphs (hundreds of nodes), embeddings for larger graphs using
these techniques are seldom structurally informative.
To this end, several approaches have been developed
for embedding graphs while preserving specific structural 
properties. 

Pioneering methods (e.g. \cite{kamada}) have resorted to 
MDS, which seeks a low-dimensional depiction of high-dimensional data in which pairwise
Euclidean distances between embedding coordinates are close (in a
well-defined sense) to the dissimilarities between the original data
points \cite{InPaMMDS}, \cite{hastie}. In this case, the vertex dissimilarity 
structure is preserved through pairwise distance metrics between
vertices. 

Principal component analysis (PCA) of the
graph adjacency matrix is advocated in \cite{luospg}, leading 
to a spectral embedding whose vertices correspond
to entries of  the leading component vectors.
The structure preserving embedding algorithm 
\cite{jebara} solves a semidefinite program with linear
topology constraints so that a nearest neighbor algorithm
can recover the graph edges from the embedding. 

Visual analytics approaches developed in \cite{piwi}
and \cite{emilio} emphasize community structures with applications
to community browsing in graphs. 
Concentric graph layouts developed in \cite{finger} and 
\cite{brandes} capture notions of node hierarchy by placing
the highest ranked nodes at the center of the embedding.

Although the 
graph embedding problem has been studied for years,
development of fast and optimal visualization algorithms 
with hierarchical constraints is challenging and 
existing methods typically resort to heuristic approaches.
The growing interest in analysis of very large networks 
has prioritized the need for effectively capturing hierarchy
over aesthetic appeal in visualization.
For instance, a hierarchy-aware visual analysis of 
a global computer network is naturally more useful 
to security experts trying to protect the most
critical nodes from a viral infection. Layouts of 
metro-transit networks that clearly show terminals 
routing the bulk of traffic convey a better picture
about the most critical nodes in the event of a terrorist
attack. 

In this paper, hierarchy is captured through 
well-defined measures of node importance
collectively known as \emph{centrality} in the 
network science community (see e.g., ~\cite{sabidussi},
\cite{freeman}). 
For instance, betweenness centrality describes
the extent to which information is routed through a specific node by
measuring the fraction of all shortest paths traversing it;
see e.g.,~\cite[p. 89]{netsci}.
Other measures include closeness,
eigenvalue, and Markov centrality.
Motivated by the desire for more effective
approaches to capture centrality structure, 
dimensionality reduction techniques are 
leveraged to develop graph embedding algorithms
that adhere to hierarchy.

\subsection{Contributions}

The present paper incorporates 
centrality structure by explicitly constraining MDS and LLE
embedding criteria and developing efficient algorithms
that yield structurally informative visualizations.
First, an MDS (so-termed stress~\cite[Chap.~3]{InPaMMDS}) optimization 
criterion with radial constraints that place nodes of higher 
centrality closer to the origin of the 
graph embedding is adopted. The novel approach exploits 
the block separability inherent to the
proposed model and introduces successive approximations to
determine the optimal embedding via coordinate descent iterations. In
addition, edge crossings are minimized by regularizing the stress cost with a 
smoothness promoting term weighted by a tuning parameter.

The second approach develops a centrality-constrained
LLE visualization algorithm. Closed-form solutions for the
weight determination step and low-complexity coordinate 
descent iterations for the dimensionality
reduction step are developed yielding a fast algorithm
that scales linearly with the graph size. Moreover, the LLE approach 
inherently preserves local structure under centrality constraints
eliminating the need for a smoothness regularization penalty.

\subsection{Related work}

Although the graph embedding problem has generally been studied for
some time, the use of centralities to guide 
visualization is recent and has been addressed predominantly
in the context of scale-free networks. The derivatives of centrality
metrics \cite{correa} are used to visualize social networks
in a way that captures the sensitivity of nodes to changes 
in degree of a given node. Edge filtering based on betweenness centrality
is used by \cite{girvan} to improve the layout of large graphs. Similar 
filtering approaches in \cite{jia} and \cite{ham} remove edges
with high betweenness centrality leading to clearer layouts that capture
the structure of the network.

Closer to the objectives of the present paper is work involving radial visualizations 
\cite{brandes} generated by iteratively solving 
an MDS stress function with centrality-based weights.
However, the proposed algorithm is sensitive to initialization and offers 
no convergence guarantee, a limitation that is addressed in this paper. 
In another study whose results
have been used extensively for internet visualization, the $k$-core decomposition is 
used to hierarchically place nodes within concentric ``onion-like'' shells 
\cite{finger}. Although effective for large-scale networks, it is a heuristic 
method with no optimality associated with it.

\subsection{Structure of the paper}

The rest of this paper is organized as follows. Section~\ref{preliminaries} 
states the graph embedding problem and Section~\ref{ccmds} casts it as
MDS with centrality constraints. BCD iterations are used to solve the
MDS problem in Section~\ref{bcd} and a smoothness penalty is included
in Section \ref{bcdsmooth}. Section~\ref{cclle} solves the embedding problem
via LLE, whereas Sections~\ref{centralities} and \ref{dissimilarities} 
discuss centrality measures and node dissimilarities relevant as inputs 
to the proposed algorithms respectively. Experimental results are presented in 
Section~\ref{experiments}, and concluding remarks are given in 
Section~\ref{conclusion}.

\textit{Notation}. Upper (lower) bold face letters denote 
matrices (column vectors) and calligraphic letters
denote sets; $ \X ^T$ denotes transpose of $ \X $; 
$ \text{Tr}\left( \X \right)$ is the trace operator of matrix $\X$; 
$ \| \x \|_2$ is the Euclidean norm of $ \x $; $\e_i$ is the $i$-th 
column of the identity matrix, $\mathbf{I}$. $\X_{\mathcal{I},\mathcal{J}}$ is the 
submatrix of $\X$ obtained by retaining the common elements in rows
indexed by the elements of $\mathcal{I}$ and columns indexed by $\mathcal{J}$.
Lastly, $\partial f(.)$ denotes the subdifferential of $f(.)$.

\section{Preliminaries}
\label{preliminaries}

Consider a network represented by a graph $\mathcal{G} =
(\mathcal{V}, \mathcal{E})$, where $\mathcal{E}$ denotes the set of
edges, and $\mathcal{V}$ the set of vertices with cardinality
$|\mathcal{V}| = N$. It is assumed that $\mathcal{G}$ is
undirected, unweighted and has no loops or multi-edges between 
node pairs. Let $\delta_{ij}$ denote the real-valued
pairwise dissimilarity between two nodes $i$ and $j$
computed via a well-defined criterion. The dissimilarities
are assumed to satisfy the following properties: (i) $\delta_{ij} \ge 0$; 
(ii) $\delta_{ij} = \delta_{ji}$; and, (iii) $ \delta_{ii} = 0 $.
Consequently, it suffices to know 
the set $ \mathcal{D} := \{ \{ \delta_{ij} \}_{j=1}^N \}_{i=j+1}^N$
with cardinality $|\mathcal{D}| = N(N-1)/2$. Let 
$c_i$ denote a centrality measure assigned to node $i$, and consider the set
$\mathcal{C} := \{c_i\}_{i=1}^N$ with cardinality $|\mathcal{C}| = N$.
Node centralities can be obtained using a number 
of algorithms~\cite[Chap. 4]{netsci}.

Given $\mathcal{G}$, $\mathcal{D}$, $\mathcal{C}$,
and the prescribed embedding dimension $p$ (typically $p \in \{2, 3\}$),
 the graph embedding task amounts to finding
the set of $p \times 1$ vectors $ \mathcal{X} := \left\lbrace \x_i 
\right\rbrace_{i=1}^N $ which ``respect'' the network structure 
characteristics encoded through $\mathcal{D}$ and $\mathcal{C}$. 
The present paper puts forth two approaches, based upon
dimensionality reduction, which ensure that $\mathcal{X}$ 
adheres to these constraints from different but related 
viewpoints. 

\section{Centrality-constrained MDS}
\label{ccmds}

MDS amounts to finding
vectors $\left\lbrace \x_i \right\rbrace_{i=1}^N $ so that the
embedding coordinates $\x_i$ and $\x_j$ satisfy $ \| \x_i - \x_j\|_2 \approx \delta_{ij}$ by
solving the following problem:
\begin{equation}
\label{eq1} (\text{P}0)\;\;\;\; \hat{\mathcal{X}} =  \underset{\x_1, \dots,
\x_N}{\operatorname{arg\,min}} \text{ }\frac{1}{2} \sum_{i =
1}^{N}\sum_{j = 1}^{N}[\left\| \mathbf{x}_i - \mathbf{x}_j
\right\|_2 - \delta_{ij}]^2.
\end{equation}

Centrality structure will be imposed on \eqref{eq1} by constraining
$\x_i$ to have a centrality-dependent radial distance $f(c_i)$,
where $f(.)$ is a monotone decreasing function. For instance,
$f(c_i) = \alpha e^{-\beta c_i}$ where $\alpha$ and $\beta$ are positive constants.
The resulting constrained optimization problem now becomes
\begin{eqnarray}
\label{eq2} \nonumber (\text{P}1)\;\;\;\;  \hat{\mathcal{X}}
 = \underset{\x_1, \dots,
\x_N}{\operatorname{arg\,min}} &
  \frac{1}{2}
  \sum\limits_{i = 1}^{N}\sum\limits_{j = 1}^{N}
  \left[
   \left\| \mathbf{x}_i - \mathbf{x}_j \right\|_2 - \delta_{ij}
  \right]^2 \\
 \text{s. to} &  \left\| \mathbf{x}_i \right\|_2 = f(c_i),\text{ }i = 1,\dots,N.
\end{eqnarray}

Although $\text{P}0$ is non-convex, standard solvers rely on
gradient descent iterations but have no guarantees of convergence to
the global optima \cite{buja}. Lack of convexity is exacerbated in
$\text{P}1$ by the non-convex constraint set rendering its solution
even more challenging than that of $\text{P}0$. However, considering
a single embedding vector $\x_i$, and fixing the rest $\{\x_j\}_{j
\ne i}$, the constraint set simplifies to $\| \x_i \|_2 = f(c_i)$,
for which an appropriate relaxation can be sought. Key to the
algorithm proposed next lies in this inherent decoupling of the
centrality constraints.
\begin{remark}
\label{remark1}
A number of stress costs
with well-documented merits have been reported for MDS.
The choice made in $P0$ is motivated
by convenience of adaptation to BCD iterations discussed in
Section \ref{bcd}. Moreover, more general forms include weights,
which have been ignored without loss of generality.
\end{remark}

\section{BCD successive approximations}
\label{bcd}

By exploiting the separable nature of the cost as well as the norm
constraints in \eqref{eq2}, BCD will be
developed in this section to arrive at a solution approaching the
global optimum. To this end, the centering constraint $\sum_{i=1}^N
\x_i = {\bf 0}$, typically invoked to fix the inherent translation
ambiguity, will be dropped first so that the problem remains
decoupled across nodes. The effect of this relaxation can be
compensated for by computing the centroid of the solution of
\eqref{eq2}, and subtracting it from each coordinate. The $N$
equality norm constraints are also relaxed to $\| \x_i \|_2 \le
f(c_i)$. Although the entire constraint set is non-convex, each
relaxed constraint is a convex and closed Euclidean ball with
respect to each node in the network.

Let $\x_i^r$ denote the minimizer of the optimization problem over
block $i$, when the remaining blocks $\left\lbrace
\x_j\right\rbrace_{j \ne i}$ are fixed during the BCD iteration $r$.
By fixing the blocks $\left\lbrace \x_j \right\rbrace_{j \ne i}$ to
their most recently updated values, the sought embedding
is obtained as
\begin{eqnarray}
\label{eq3}
\nonumber
\hat{\x}_i & = &
\underset{ \x }{\operatorname{arg\,min}}
  \frac{1}{2}
  \sum_{j \ne i}
  \left[
   \left\| \mathbf{x} - \mathbf{x}_j \right\|_2 - \delta_{ij}
  \right]^2 \\
\text{s. to}& & \left\| \mathbf{x} \right\|_2
\le f(c_i)
\end{eqnarray}
or equivalently as
\begin{eqnarray}
\label{eq4}
\nonumber
\underset{ \x }{\arg\min} & &
\frac{(N-1)}{2} \| \x \|_2^2 - \x^T( \sum_{j < i} \x_j^r + \sum_{j > i} \x_j^{r-1}) \\
\nonumber
& & - \sum_{j < i} \delta_{ij} \| \x - \x_j^r \|_2 - \sum_{j > i} \delta_{ij} \| \x - \x_j^{r-1} \|_2 \\
\text{s. to} & & \left\| \x \right\|_2 \le f(c_i)
\end{eqnarray}
where $ \sum_{j < i} (.) := \sum_{j = 1}^{i-1} (.)$
and $ \sum_{j > i} (.) := \sum_{j = i+1}^{N} (.)$.
With the last two sums in the cost function of \eqref{eq4} being non-convex and
non-smooth, convergence of the BCD algorithm cannot be guaranteed
\cite[p. 272]{bertsekas}. Moreover, it is desired to have each per-iteration
subproblem solvable to global optimality, in closed form and at a
minimum computational cost. The proposed approach seeks a global
upper bound of the objective with the desirable properties of
smoothness and convexity. To this end, consider the function $\Psi(
\x) := \psi_1(\x) - \psi_2 (\x)$, where
\begin{equation}
\label{eq4a} \psi_1(\x) := \frac{(N-1)}{2}\| \x \|_2^2 - \x^T(
\sum\limits_{j < i} \x_j^r + \sum\limits_{j > i} \x_j^{r-1})
\end{equation}
and
\begin{equation}
\label{eq4b} \psi_2 (\x) :=  \sum\limits_{j < i} \delta_{ij} \|
\x - \x_j^r \|_2 + \sum\limits_{j > i} \delta_{ij} \| \x -
\x_j^{r-1} \|_2.
\end{equation}

Note that $ \psi_1(\x) $ is a convex quadratic function, and that
$\psi_2(\x) $ is convex (with respect to $\x$) but
non-differentiable. The first-order approximation of \eqref{eq4b} at
any point in its domain is a global under-estimate of $\psi_2(\x)
$. Despite the non-smoothness at some points, such a lower bound can
always be established using its subdifferential. As a consequence of
the convexity of $\psi_2(\x) $, it holds that~\cite[p. 731]{bertsekas}
\begin{equation}
\label{eq5}
\psi_2(\x)  \ge  \psi_2(\x_0) +
\mathbf{g}^T(\x_0)(\x - \x_0),  \forall \x \in \text{dom}(\psi_2)
\end{equation}
where $\mathbf{g}(\x) \in \partial \psi_2(\x)$ is a subgradient
within the subdifferential set, $\partial \psi_2(\x)$ of $
\psi_2(\x)$. The subdifferential of $\| \x - \x_j \|_2$ with
respect to $\x$ is given by
\begin{equation}
\label{eq6}
\partial_{\x} \| \x - \x_j \|_2 =
\begin{cases}
\frac{\x - \x_j}{\| \x - \x_j \|_2}, & \text{ if } \x \ne \x_j \\
\mathbf{s} \in \mathbb{R}^p : \text{ } \| \mathbf{s} \|_2 \le 1, & \text{ otherwise }
\end{cases}
\end{equation}
which implies that
\begin{equation}
\label{eq7}
\partial_{\x} \psi_2(\x) = \sum_{j = 1}^N \delta_{ij} \partial_{\x} \| \x - \x_j \|_2.
\end{equation}
Using \eqref{eq5}, it is possible to lower bound \eqref{eq4b} by
\begin{eqnarray}
\label{eq8}
\nonumber
\psi_2'(\x, \x_0) = \sum\limits_{j < i} \delta_{ij} \left[ \| \x_0 - \x_j^r \|_2 +
 (\mathbf{g}_j^{r})^T(\x_0)(\x - \x_0) \right]  \\
  + \sum\limits_{j > i} \delta_{ij} \left[  \| \x_0 - \x_j^{r-1} \|_2
+ (\mathbf{g}_j^{r-1})^T(\x_0)(\x - \x_0)\right].
\end{eqnarray}
Consider now $\Phi(\x, \x_0) := \psi_1(\x) - \psi_2'(\x,
\x_0)$, and note that $\Phi(\x, \x_0)$ is convex and globally upper bounds
the cost in \eqref{eq4}. The proposed BCD algorithm
involves successive approximations using \eqref{eq8}, and yields the
following QCQP per block
\begin{equation}
\label{eq9} (\text{P}2) \quad\quad \underset{\left\lbrace \x :
\left\| \mathbf{x} \right\|_2 \le f(c_i)
\right\rbrace}{\operatorname{arg\,min}}  \Phi(\x, \x_0).
\end{equation}
For convergence, $\x_0$ must be selected to satisfy the following
conditions \cite{meis}:
\begin{subequations}
\begin{eqnarray}
\label{eq10}
\Phi(\x_0, \x_0) & = & \Psi(\x_0), \quad \forall \x_0 \in \mathcal{F} \\
\label{eq10b} \Phi(\x, \x_0) & \ge & \Psi(\x), \quad \| \x
\|_2 \le f(c_i),  \forall i
\end{eqnarray}
\end{subequations}
where $\mathcal{F} := \bigcup_{i = 1}^{N} \left\lbrace \x : \|
\x \|_2 \le f(c_i)\right\rbrace $. In addition, $\Phi(\x, \x_0)$
must be continuous in $(\x, \x_0)$. 
\begin{proposition}
\label{prop1}
The conditions for convergence in \eqref{eq10} and
\eqref{eq10b} are satisfied by selecting $\x_0 =
\x^{r-1}$.
\end{proposition}
The proof of Proposition \ref{prop1} involves substituting $\x_0 =
\x^{r-1}$ in \eqref{eq10} and \eqref{eq10b} (see Appendix \ref{appendix:prop1}).
Taking successive approximations around $\x^{r-1}$
in $\text{P}2$, ensures the uniqueness of
\begin{eqnarray}
\label{eq11}
\nonumber
\x_i^r &=& \underset{\left\lbrace \x : \left\| \mathbf{x} \right\|_2
\le f(c_i)  \right\rbrace}{\operatorname{arg\,min}}  \frac{(N-1)}{2} \x^T\x \\
\nonumber
& & - \x^T \Big[ \sum\limits_{j < i}(\x_j^r + \delta_{ij}\mathbf{g}_j^r(\x^{r-1}) ) \\
& & + \sum\limits_{j > i}(\x_j^{r-1} + \delta_{ij}\mathbf{g}_j^{r-1}(\x^{r-1}) ) \Big].
\end{eqnarray}
Solving \eqref{eq11} amounts to obtaining the solution of the
unconstrained QP, $(\x^*)^r$, and projecting it onto~$\left\lbrace
\x : \left\| \mathbf{x} \right\|_2 \le f(c_i) \right\rbrace$;
that is,
\begin{equation}
\label{eq12}
\x_i^r =
\begin{cases}
\frac{(\x^*)^r}{\| (\x^*)^r \|_2} f(c_i),\;\; \text{if}\;\;\; \| (\x^*)^r \|_2 > f(c_i) \\
(\x^*)^r , \text{ otherwise }
\end{cases}
\end{equation}
where
\setlength{\abovedisplayskip}{3pt}
\setlength{\belowdisplayskip}{3pt}
\begin{eqnarray}
\label{eq13}
\nonumber
(\x^*)^r & =& \frac{1}{N-1} \Bigg[ \sum\limits_{j < i}(\x_j^r + \delta_{ij}\mathbf{g}_j^r(\x^{r-1}) ) \\
& & + \sum\limits_{j > i}(\x_j^{r-1} + \delta_{ij}\mathbf{g}_j^{r-1}(\x^{r-1}) ) \Bigg].
\end{eqnarray}
It is desirable but not necessary that the algorithm converges
because depending on the application, reasonable network
visualizations can be found with fewer iterations. In fact,
successive approximations merely provide a more refined graph
embedding that maybe more aesthetically appealing.

Although the proposed algorithm is guaranteed to converge, the
solution is only unique up to a rotation and a translation (cf.
MDS). In order to eliminate the translational ambiguity, the
embedding can be centered at the origin. Assuming that the optimal
blocks determined within outer iteration $r$ are reassembled into
the embedding matrix $\X^r := \left[ (\x_1^r)^T, \dots, (\x_N^r)^T \right]^T$,
the final step involves subtracting the mean from each coordinate
using the centering operator as follows, $\X = (\mathbf{I} -
N^{-1}\ones \ones^T)\X^r$, where $\mathbf{I}$ denotes the $N \times
N$ identity matrix, and $\ones$ is the $N \times 1$ vector of all
ones.

The novel centrality-constrained MDS (CC-MDS) 
graph embedding scheme is summarized as Algorithm
\ref{alg1} with matrix $\mathbf{\Delta}$ having $(i,j)$th entry the
dissimilarity $\delta_{ij}$, and $\epsilon$ denoting a tolerance level.
\begin{algorithm}
    \caption{CC-MDS}
\label{alg1}
\begin{algorithmic}[1]
   \STATE {\bfseries Input:}  $\left\lbrace c_i \right\rbrace_{i=1}^N$,  $\mathbf{\Delta}$, $\epsilon$
   \STATE Initialize $\X^0$, $r=0$
   \REPEAT
   \STATE $r = r + 1$
   \FOR{$i=1 \dots N$}
   \STATE Compute $\x_i^r$ according to \eqref{eq12} and \eqref{eq13}
   \STATE $\X^r(i,:) = (\x_i^r)^T$
   \ENDFOR
   \UNTIL{$\| \X^{r} - \X^{r-1} \|_F \le \epsilon$}
   \STATE $\X = (\mathbf{I} - \frac{1}{N}\ones \ones^T)\X^r$
\end{algorithmic}
\end{algorithm}
\vspace*{-0.5cm}
\section{Enforcing graph smoothness}
\label{bcdsmooth}
In this section, the MDS stress in
\eqref{eq3} is regularized through an additional constraint that
encourages smoothness over the graph. Intuitively, despite the
requirement that the node placement in low-dimensional Euclidean
space respects inherent network structure, through preserving e.g.,
node centralities, neighboring nodes in a graph-theoretic sense
(meaning nodes that share an edge) are expected to be close in
Euclidean distance within the embedding. An example of an application
where smoothness is well motivated is the visualization of 
transportation networks that are defined over geographical regions.
Despite the interest in their centrality structure, embeddings that
place nodes that are both geographically close and adjacent in 
the representative graph are more acceptable to users that are
familiar with their conventional (and often geographically
representative) maps.
Such a requirement can be
captured by incorporating a constraint that discourages large
distances between neighboring nodes. In essence, this constraint
enforces smoothness over the graph embedding.

A popular choice of a smoothness-promoting function is $ h(\X) :=
\text{Tr}(\X^T\L\X)$, where $\text{Tr}(.)$ denotes the trace
operator, and $\L := \D - \A$ is the graph Laplacian with $\D$ a
diagonal matrix whose $(i,i)$th entry is the degree of node $i$, and
$\A$ the adjacency matrix. It can be shown that $h(\X) = (1/2)
\sum\limits_{i = 1}^{N} \sum\limits_{i = 1}^{N} a_{ij} \| \x_i -
\x_j \|_2^2$, where $a_{ij}$ is the $(i,j)$th entry of $\mathbf{A}$.
Motivated by penalty methods in optimization, the cost in
\eqref{eq2} will be augmented as follows
\begin{eqnarray}
\label{eq15} \nonumber (\text{P}3)\;\;\; \quad
\underset{\x_1, \dots, \x_N}{\operatorname{arg\,min}} &
  \frac{1}{2}
  \sum\limits_{i = 1}^{N}\sum\limits_{j = 1}^{N}
  \left[
   \left\| \mathbf{x}_i - \mathbf{x}_j \right\|_2 - \delta_{ij}
  \right]^2  \\
  \nonumber
  & +  \frac{\lambda}{2} \sum\limits_{i=1}^{N} \sum\limits_{j=1}^{N} a_{ij}\| \x_i - \x_j \|_2^2
  \\
\text{s. to} & \left\| \mathbf{x}_i \right\|_2 = f(c_i), i = 1,\dots,N
\end{eqnarray}
where the scalar $\lambda \ge 0$ controls the degree of smoothness.
The penalty term has a separable structure and is convex with
respect to $\x_i$. Consequently, $\text{P}3$ lies within the
framework of successive approximations required to solve each
per-iteration subproblem. Following the same relaxations and
invoking the successive upper bound approximations described
earlier, yields the following QCQP
\begin{eqnarray}
\label{eq16}
\nonumber
\x_i^r &=\underset{\left\lbrace \x : \left\| \mathbf{x} \right\|_2
\le f(c_i)  \right\rbrace}{\operatorname{arg\,min}}
\frac{(N + \lambda d_{ii} - 1)}{2} \x^T\x \\
\nonumber
&- \x^T[ \sum\limits_{j < i}((1+\lambda a_{ij})\x_j^r + \delta_{ij}\mathbf{g}_j^r(\x^{r-1}) ) \\
&+ \sum\limits_{j > i}((1+\lambda a_{ij})\x_j^{r-1} + \delta_{ij}\mathbf{g}_j^{r-1}(\x^{r-1}) ) ]
\end{eqnarray}
with $d_{ii} := \sum\limits_{j=1}^{N}a_{ij}$
denoting the degree of node $i$. \\
\indent The solution of \eqref{eq16} can be expressed as [cf.
\eqref{eq12}]
\begin{eqnarray}
\label{eq17}
\nonumber
(\x^*)^r &=\frac{1}{N + \lambda d_{ii} -1}[ \sum\limits_{j < i}((1+\lambda a_{ij})\x_j^r + \delta_{ij}\mathbf{g}_j^r(\x^{r-1}) ) \\
&+ \sum\limits_{j > i}((1+\lambda a_{ij})\x_j^{r-1} + \delta_{ij}\mathbf{g}_j^{r-1}(\x^{r-1}))].
\end{eqnarray}
With $\lambda$ given, Algorithm \ref{alg2} summarizes the steps to
determine the constrained embedding with a smoothness penalty.
\begin{algorithm}
    \caption{CC-MDS incorporating smoothness}
\label{alg2}
\begin{algorithmic}[1]
   \STATE {\bfseries Input:}  $\A$, $\left\lbrace c_i \right\rbrace_{i=1}^N$,  $\mathbf{\Delta}$, $\epsilon$, $\lambda$
   \STATE Initialize $\X^0$, $r=0$
   \REPEAT
   \STATE $r = r + 1$
   \FOR{$i=1 \dots N$}
   \STATE Compute $\x_i^r$ according to \eqref{eq12} and \eqref{eq17}
   \STATE $\X^r(i,:) = (\x_i^r)^T$
   \ENDFOR
   \UNTIL{$\| \X^{r} - \X^{r-1} \|_F \le \epsilon$}
   \STATE $\X = (\mathbf{I} - \frac{1}{N}\ones \ones^T)\X^r$
\end{algorithmic}
\end{algorithm}
\vspace{-0.3cm}

\section{Centrality-constrained LLE}
\label{cclle}

LLE belongs to a family of non-linear dimensionality reduction 
techniques which impose a low-dimensional manifold structure on 
the data with the objective of seeking an embedding that preserves 
the local structure on the manifold \cite[Chap. 14]{hastie}.
In particular, LLE 
accomplishes this by approximating each data point by a linear 
combination of its neighbors determined by a well-defined criterion 
followed by construction of a lower dimensional embedding that best 
preserves the approximations. Suppose 
$\left \lbrace \y_i \in \mathbb{R}^q \right \rbrace_{i = 1}^N$
are data points that lie close to a manifold in $\mathbb{R}^q$,
the dimensionality reduction problem seeks the vectors  
$\left \lbrace \x_i \in \mathbb{R}^p \right \rbrace_{i = 1}^N$
where $ p \ll q$. In this case, the first step determines the $K$-nearest 
neighbors for each point, $i$, denoted by $\mathcal{N}_i$. 
Then each point is approximated by a linear combination
of its neighbors by solving the following optimization problem 
\begin{eqnarray}
\label{eqlle1}
\nonumber
\underset{w_{i1},\dots,w_{iK }}{\operatorname{arg\,min}} & &
\| \y_i - \sum\limits_{j \in \mathcal{N}_i} w_{ij}\y_j \|_2^2 \\
\text{s. to} & & \sum\limits_{j \in \mathcal{N}_i} w_{ij} = 1, \;\; i = 1, \dots, N
\end{eqnarray}
where $\left \lbrace w_{ij} \right \rbrace_{j = 1}^K$ are the 
$K$ reconstruction weights for point $i$ and the constraint 
enforces shift invariance. Setting $w_{ij} = 0$ for 
$j \notin \mathcal{N}_i$, the final step determines 
$\left \lbrace \x_i \in \mathbb{R}^p \right \rbrace_{i = 1}^N$
while preserving the weights by solving 
\begin{eqnarray}
\label{eqlle2}
\nonumber
\underset{\x_1,\dots, \x_N}{\operatorname{arg\,min}} & &
\sum\limits_{i = 1}^{N} \| \x_i -  \sum\limits_{j = 1}^{N} w_{ij}\x_j  \|_2^2 \\
\text{s. to} & & \sum\limits_{i=1}^{N} \x_i = \mathbf{0}, \quad \frac{1}{N} 
\sum\limits_{i=1}^{N} \x_i \x_i^T = \mathbf{I}
\end{eqnarray}
with the equality constraints introduced to center the embedding at
the origin with a unit covariance.

\subsection{Centrality constraints}
\label{sub_lle_cc}
This section proposes an adaptation of the LLE algorithm to the 
graph embedding problem by imposing a centrality structure through 
appropriately selected constraints. Since $\mathcal{G}$ is given,
selection of neighbors for each node, $i$, is straightforward
and can be accomplished by assigning $\mathcal{N}_i$ to the set
of its $n$-hop neighbors. As a result, each node takes on a different
number of neighbors, as dictated by the topology of the graph.
 
Associating with node $i$ a vector, $\y_i$, 
of arbitrary dimension $ q \gg p$, and adding the centrality constraint 
$( \sum_{j \in \mathcal{N}_i} w_{ij}\y_j )^T( \sum_{j \in \mathcal{N}_i} w_{ij}\y_j ) = f^2(c_i) $
to \eqref{eqlle1} per node yields
\begin{eqnarray}
\label{eqlle3}
\nonumber
\text{(P4)} \;\;\;\; \mathbf{w}_i =  \underset{  \left \lbrace \mathbf{w} : \; \mathbf{1}^T\mathbf{w} = 1 \right \rbrace  }
{\operatorname{arg\,min}} & & \| \mathbf{y}_i - \mathbf{Y}_i \mathbf{w}_i \|_2^2 \\
\text{s. to} & & \| \mathbf{Y}_i\mathbf{w}_i \|_2^2= f^2(c_i)
\end{eqnarray}

\noindent
where $\mathbf{Y}_i := [\mathbf{y}_1^i, \dots, \mathbf{y}_K^i]$ contains the $K$
$n$-hop neighbors of $i$, $\left \lbrace \y_j^i  \right \rbrace_{j = 1}^K$, and 
$\mathbf{w}_i := [ w_{i1}, \dots, w_{iK} ]^T$. Similarly, the final LLE step
is modified as follows

\begin{eqnarray}
\label{eqlle4}
\nonumber
\text{(P5)} \;\;\;\; \underset{ \mathbf{x}_1, \dots, \mathbf{x}_N }{\operatorname{arg\,min}} & & \sum\limits_{i=1}^{N}\| \mathbf{x}_i - \sum\limits_{j=1}^{N}w_{ij}\mathbf{x}_j \|_2^2 \\
\text{s. to} & &  \|\mathbf{x}_i \|_2^2 = f^2(c_i), \text{ }i = 1,\dots,N
\end{eqnarray}

\noindent
with the $\mathbf{0}$-mean constraint compensated for by a centering 
operation after the optimal vectors $\left \lbrace \x_i \right \rbrace_{i=1}^N$
have been determined. P4 is nonconvex due to the inclusion
of a quadratic equality constraint. Relaxing the constraint to the inequality
$( \sum_{j \in \mathcal{N}_i} w_{ij}\y_j )^T( \sum_{j \in \mathcal{N}_i} w_{ij}\y_j )
\le f^2(c_i)$ leads to a convex problem which can be easily solved.

In general, $\left \lbrace \y_i \right \rbrace_{i = 1}^N$ are unknown and 
a graph embedding must be determined entirely from $\mathcal{G}$. However, 
the terms in both the cost function and the constraint in \eqref{eqlle3} 
are inner products, $ \y_i^T\y_j$ for all $i, j \in \left \lbrace 1, \dots, N
\right \rbrace$, which can be approximated using the dissimilarities
$\{ \{\delta_{ij} \}_{j=1}^N \}_{i=j+1}^N$. This is reminiscent of classical
MDS which seeks vectors $\left \lbrace \y_i \right \rbrace_{i = 1}^N$ 
so that $\| \y_i - \y_j \|_2 \approx \delta_{ij} $ for any pair
of points $i$ and $j$. To this end, \eqref{eqlle3} can be written as

\begin{eqnarray}
\label{eqlle5}
\nonumber
\w_i = \underset{ \left \lbrace \w : \; \mathbf{1}^T\w = 1 \right \rbrace }
{\operatorname{arg\,min}} & & \w^T \Y_i^T \Y_i \w - 2 \y_i^T \Y_i \w \\
\text{s. to} & & \w^T \Y_i^T \Y_i \w \le f^2(c_i)
\end{eqnarray}

\noindent
If $ \bm{\mathcal{ D}} $ denotes the $ N \times N $ matrix whose $ (i, j)$th entry is the square
Euclidean distance between $\y_i$ and $\y_j$ i.e., $ \left[ \bm{\mathcal{ D}} \right]_{ij} 
:= \| \y_i - \y_j \|_2^2$, and
$\Y := [\y_1, \dots, \y_N]$, 
then

\begin{equation}
\label{eqlle6}
\bm{\mathcal{ D}} = \boldsymbol{\psi}\mathbf{1}^T + \mathbf{1}\boldsymbol{\psi}^T - 2\Y^T\Y
\end{equation}

\noindent
where $\boldsymbol{\psi} \in \mathbb{R}^N $ has entries $\psi_i = \| \y_i \|_2^2$.
Denoting the centering matrix as $ \mathbf{J} := \mathbf{I} - N^{-1}\mathbf{11}^T$, 
and applying the double-centering operation to $ \bm{\mathcal{ D}} $ yields
\begin{eqnarray}
\label{eqlle7}
\nonumber
-\frac{1}{2}\mathbf{J} \bm{\mathcal{ D}} \mathbf{J} &=& -\frac{1}{2}\mathbf{J}\boldsymbol{\psi}\mathbf{1}^T\mathbf{J} - \frac{1}{2}\mathbf{J}\mathbf{1}\boldsymbol{\psi}^T\mathbf{J} + \mathbf{J}\Y^T \Y\mathbf{J} \\
& = & \Y^T \Y
\end{eqnarray}
which is the inner product matrix of the vectors $\left \lbrace \y_i \right \rbrace_{i=1}^N$, i.e., 
$ [\Y^T\Y]_{ij} = \y_i^T\y_j$, since $\Y^T \Y$ is already double-centered.
Although $ \bm{\mathcal{ D}} $ is unknown, it is possible to estimate $\Y^T \Y \approx 
-\frac{1}{2}\mathbf{J}\mathbf{\Delta}^{\left( 2 \right)}\mathbf{J}$ from \eqref{eqlle7}
using square dissimilarity measurements
$ [\mathbf{\Delta}^{\left( 2 \right)}]_{ij} = \delta_{ij}^2$ 
as surrogates to the square Euclidean distances. 
Since $\mathbf{\Delta}^{\left( 2 \right)}$ can be determined 
from the graph topology alone using graph-theoretic distance measures
(e.g., shortest path distances), \eqref{eqlle5} can be determined
in closed form as shown next.
If $\mathbf{H} := -\frac{1}{2}\mathbf{J}\mathbf{\Delta}^{\left( 2 \right)}\mathbf{J} $,
and $\mathbf{H}_i$ denotes the estimate of $ \Y_i^T \Y_i $, then 
$\mathbf{H}_i = \mathbf{H}_{\mathcal{N}_i, \mathcal{N}_i}$. Similarly,
the estimate of $ \Y_i^T \y_i $ is $\mathbf{h}_i = \mathbf{H}_{\mathcal{N}_i, i} $.
Using this notation, the first step of LLE in the proposed approach
seeks the solution to the convex constrained QP
\begin{eqnarray}
\label{eqlle8}
\nonumber
\text{(P6)} \;\; \w_i =  \underset{ \left \lbrace \w : \; \mathbf{1}^T\w = 1
\right \rbrace }{\operatorname{arg\,min}} & \hspace{-2.0cm} \w^T \mathbf{H}_i \w - 2 \mathbf{h}_i^T \w \\
\nonumber
\text{s. to} &  \w^T \mathbf{H}_i \w \le f^2(c_i), i = 1, \dots, N \\
& & .
\end{eqnarray}
\begin{remark}
\label{remark2}
Since the entries of $\mathbf{H}$ are inner products,
it is a kernel matrix and can be replaced by a graph kernel
that reliably captures similarities between nodes in 
$\mathcal{G}$. The Laplacian pseudoinverse, $\mathbf{L}^{\dagger}$
is such a kernel in the space spanned by the graph nodes, where 
node $i$ is represented by $\mathbf{e}_i$ \cite{fouss}. In fact,
$(-1/2) \mathbf{J \Delta }^{(2)} \mathbf{J}$ and $\mathbf{L}^{\dagger}$ are equivalent 
when the entries of $\mathbf{\Delta}^{(2)}$ are average commute times,
briefly discussed in Section \ref{dissimilarities}.
\end{remark}

In order to solve P6 per node, Lagrange multipliers $\gamma$ 
and $\mu$ corresponding to the inequality and equality constraints are introduced  
and lead to the Lagrangian
\begin{eqnarray}
\label{eqlle9}
\nonumber
L(\w, \gamma, \mu)  &=&  \w^T \H_i \w - 2\h_i^T\w - \gamma(\w^T \H_i \w - f^2(c_i)) \\
& & + \mu (\ones^T\w - 1).
\end{eqnarray}
Assuming Slater's condition is satisfied \cite{BoVa04}, a zero duality gap is achieved if the following KKT conditions are satisfied by the optimal primal and dual variables:
\begin{subequations}
\begin{eqnarray}
\label{eqlle10}
\w^T \H_i \w - f^2(c_i) & \le & 0 \quad  \label{eqlle10a}\\
\ones^T\w - 1 & = & 0 \quad  \label{eqlle10b}\\
\gamma & \ge & 0 \quad  \label{eqlle10c}\\
\gamma ( \w^T \H_i \w - f^2(c_i) ) & = & 0 \quad   \label{eqlle10d}\\
\nabla L(\w, \gamma, \mu) = (1+\gamma)\H_i \w + \frac{\mu}{2}\ones - \h_i & = & 0. \quad  \label{eqlle10e}
\end{eqnarray}
\end{subequations}
Upon solving the KKT conditions (see Appendix 
\ref{appendix:kkt} for derivation) it follows
that

\begin{eqnarray}
\label{eqlle11}
\nonumber
\mu^* &=& \hspace{-0.2cm} \frac{2(\ones^T\H_i^{-1}\h_i)}{\ones^T\H_i^{-1}\ones}   \\
       & & \hspace{-0.2cm} + 2\left\lbrace \frac{\h_i^T\H_i^{-1}(\ones\ones^T\H_i^{-1}\h_i - \h_i\ones^T\H_i^{-1}\ones)}{(\ones^T\H_i^{-1}\ones)^2 - (\ones^T\H_i^{-1}\ones)^3 f^2(c_i)} 
       \right \rbrace^{\frac{1}{2}} 
\end{eqnarray}

\noindent
and

\begin{equation}
\label{eqlle12}
\hspace{-2.0cm} \gamma^* =  \ones^T\H_i^{-1}\h_i - \frac{\mu^*}{2}\ones^T\H_i^{-1}\ones - 1.
\end{equation}

\noindent
The optimal value, $\w_i^*$, is determined as follows

\begin{equation}
\label{eqlle13}
\w_i^* = \left\{ 
  \begin{array}{l l}
    \frac{1}{(1+\gamma^*)} \H_i^{-1}\left(\h_i - \frac{\mu^*}{2} \ones \right), 
    & \quad \text{if $\gamma^* \in \mathbb{R}_+$}\\
    \H_i^{-1}\left(\h_i - \frac{\ones^T\H_i^{-1}\h_i - 1}{\ones^T\H_i^{-1}\ones} 
    \ones \right), & \quad \text{otherwise}.\\
  \end{array} \right.
\end{equation}
Although $\mathbf{H}_i$ is provably positive semidefinite, it is not guaranteed
to be non-singular as required by \eqref{eqlle11} - \eqref{eqlle13}. Since
$\mathbf{H}_i$ is an approximation to inner products, ensuring 
positive definiteness by uniformly perturbing the diagonal entries slightly is reasonable
in this case, i.e., $\mathbf{H}_i \gets \mathbf{H}_i + \sigma \mathbf{I}_K$,
 where $\sigma$ is a small number. 

Setting all $w_{ij} = 0$ for $j \notin \mathcal{N}_i$,  the sought graph embedding
is determined by solving P5  over $ \left\lbrace \x_i  \right\rbrace_{i=1}^N $.
P5 is non-convex and global optimality is not guaranteed. However,
the problem decouples over vectors  $ \left\lbrace \x_i  \right\rbrace_{i=1}^N $
and this block separability can be exploited to solve it via BCD iterations. 
Inner iteration $i$ under outer iteration $r$ involves solving

\begin{eqnarray}
\label{eqlle14}
\nonumber
\x_i^r  =  
\underset{\x}{\operatorname{arg\,min}} & & \| \x -
 \sum\limits_{j<r} w_{ij}\x_j^r - \sum\limits_{j>r} w_{ij}\x_j^{r - 1}  \|_2^2 \\
\text{s. to} & & \| \x \|_2^2 = f^2(c_i).
\end{eqnarray}

\noindent
With $\mathbf{v}_i^r := \sum_{j<r} w_{ij}\x_j^r + \sum_{j>r} w_{ij}\x_j^{r - 1} $ 
and $\nu$ denoting a Lagrange multiplier, and minimization
of the Lagrangian for \eqref{eqlle14}, leads to
\begin{equation}
\label{eqlle15}
\x_i^r = \underset{\x}{\operatorname{arg\,min}} \quad \| \x -
 \mathbf{v}_i^r \|_2^2 + \nu ( \| \x \|_2^2  - f^2(c_i))
\end{equation}
which yields
\begin{equation}
\label{eqlle16}
\x_i^r = \frac{\mathbf{v}_i^r}{1+\nu}.
\end{equation}
Substituting \eqref{eqlle16} into the equality constraint 
in \eqref{eqlle14} leads to the closed-form per-iteration update of $\x_i$

\begin{equation}
\label{eq275}
\x_i^r =
\begin{cases}
\frac{ \mathbf{v}_i^r}{ \| \mathbf{v}_i^r \|_2 }f(c_i), \quad \text{ if } \| \mathbf{v}_i^r \|_2 > 0 \\
\x_i^{r - 1}, \quad \text{otherwise}.
\end{cases}
\end{equation}

\noindent
Letting $\X^r$ denote the embedding matrix after $r$ BCD iterations, 
the operation $\X = (\mathbf{I} - N^{-1}\mathbf{11}^T)\X^r$ centers 
$\left\lbrace \x_i^r \right\rbrace_{i=1}^N$ to the 
origin in order to satisfy the shift invariance property of the embedding.

Algorithm \ref{alg3} summarizes the steps outlined in this section
for the centrality-constrained LLE (CC-LLE) graph embedding approach. It is assumed that
the only inputs to the algorithm are the graph topology $\mathcal{G}$,
the centrality measures, $\left\lbrace c_i\right\rbrace_{i=1}^{N}$,
the graph embedding dimension $p$, the square dissimilarity
matrix $\mathbf{\Delta}^{(2)}$, and the number of hops to consider
for neighborhood selection per node, $n$.

\begin{algorithm}
    \caption{Graph embedding via CC-LLE}
\label{alg3}
\begin{algorithmic}[1]
   \STATE {\bfseries Input:}  $\mathcal{G}$, $\left\lbrace c_i \right\rbrace_{i=1}^N$,  $\mathbf{\Delta}^{(2)}$, $\epsilon$, $n$, $p$
   \STATE Set $\mathbf{H} = -\frac{1}{2}\mathbf{J}\mathbf{\Delta}^{(2)}\mathbf{J}$
   \FOR{$i=1 \dots N$}
   \STATE Set $\mathcal{N}_i$ to $n$-hop neighbors of $i$
   \STATE $\mathbf{H}_i = \mathbf{H}_{\mathcal{N}_i, \mathcal{N}_i} $, 
          $\mathbf{h}_i = \mathbf{H}_{\mathcal{N}_i, i}$
   \STATE Compute $\mu$, $\gamma$ from \eqref{eqlle11}, \eqref{eqlle12}
   \STATE Compute $\w_i$ from \eqref{eqlle13}
   \STATE Set $w_{ij} = 0$ for $j \notin \mathcal{N}_i$
   \ENDFOR
   \STATE Initialize $\X^0$, $r=0$
   \REPEAT
   \STATE $r = r + 1$
   \FOR{$i=1 \dots N$}
   \STATE Compute $\x_i^r$ according to \eqref{eq275}
   \STATE $\X^r(i,:) = (\x_i^r)^T$
   \ENDFOR
   \UNTIL{$\| \X^{r} - \X^{r-1} \|_F \le \epsilon$}
   \STATE $\X = (\mathbf{I} - \frac{1}{N}\ones \ones^T)\X^r$
\end{algorithmic}
\end{algorithm}
\vspace{-0.3cm}

\section{Network Centrality Measures}
\label{centralities}
Centrality measures offer a valuable means of quantifying the level of importance
attributed to a specific node or edge within the network. For instance, such 
measures address questions pertaining to the most authoritative
authors in a research community, the most influential web pages, or which genes would 
be most lethal to an organism if deleted. Such questions often arise in
the context of social, information and biological networks and 
a plethora of centrality measures ranging 
from straightforward ones like the node degree to the more sophisticated ones
like the PageRank \cite{pagerank} have been proposed in the network science 
literature.
Although the approaches proposed in this paper presume that $\{ c_i \}_{i=1}^N$ are known beforehand, 
this section briefly discusses two of the common centrality measures that will be 
referred to in a later section on numerical experiments.

Closeness centrality captures the extent to which a particular node is close to many other
nodes in the network. Although several methods are available, the standard approach
determines the inverse of the total geodesic distance between the node whose closeness centrality
is desired and all the other nodes in the network \cite{sabidussi}. If $c_i^{cl}$ denotes 
the closeness centrality
of node $i$, then
\begin{equation}
\label{eq_cent_1}
c_i^{cl} := \frac{1}{\sum_{j \in \mathcal{V}} d_{ij}}
\end{equation}
where $d_{ij}$ is the geodesic distance (lowest sum of edge weights) between nodes $i$ and $j$.
In the context of visualization requirements for large social networks, an embedding which encodes the level
of influence of a paper in a citation network or an autonomous system in the internet
would set $\{ c_i \}_{i=1}^N$ to be the closeness centralities.

Betweenness centrality, on the other hand, summarizes the extent to which nodes 
are located between other pairs of nodes. It measures the fraction of shortest paths
that traverse a node to all shortest paths in the network and is commonly defined as  
\begin{equation}
\label{eq_cent_2}
c_i^b := \frac{\sum_{j \ne k \ne i \in \mathcal{V}}
\sigma_{j,k}^{i}} { \sum_{i \in \mathcal{V}}\sigma_{j,k}^{i}}
\end{equation}
where $\sigma_{j,k}^{i}$ is the number of shortest paths between nodes 
$j$ and $k$ through node $i$~\cite{freeman}. An immediate application
of betweenness centrality to network visualization is envisaged in the context 
of exploratory analysis of transport networks (for instance a network whose nodes
are transit and terminal stations and edges represent connections by rail lines).
In such a network, a knowledge of stations that route the most traffic
is critical for transit planning and visualizations that effectively capture 
the betweenness centrality are well motivated.

\section{Node Dissimilarities}
\label{dissimilarities}
This section briefly discusses some of the dissimilarity measurements 
for the graph embedding task. Although $\mathcal{V}$ can represent
a plethora of inter-related objects with well defined criteria for pairwise dissimilarities
(e.g., $\delta_{ij} = \| \y_i - \y_j \|_2$, where vectors $\y_i$ 
and $\y_j$ corresponding to nodes $i$ and $j$
are available in Euclidean space),
$\mathcal{D}$ must be determined entirely from $\mathcal{G}$ in order 
to obtain embeddings that reliably encode the underlying network topology.
Node dissimilarity metrics can be broadly classified 
as: i) methods based on the number of shared neighbors, and ii)
methods based on graph-theoretic distances.

\subsection{Shared neighbors}
Graph partitioning techniques tend to favor dissimilarities based on the 
number of shared neighbors \cite{netsci}. Consider for instance
\begin{equation}
\label{eqdiss1}
\delta_{ij} = \frac{| \mathcal{N}_i \triangle \mathcal{N}_j |}{d_{(|V|)} + d_{(|V| - 1)}}
\end{equation}
where $\mathcal{A} \triangle \mathcal{B}$ denotes
the symmetric difference between sets $\mathcal{A}$ and $\mathcal{B}$; and $d_{(i)}$
denotes the $i$-th smallest element in the degree sequence of $\mathcal{G}$.
The ratio in \eqref{eqdiss1} yields a normalized metric over $[0, 1]$, and computes the number 
of single-hop neighbors of $i$ and $j$ that are not shared. 
An alternative
measure commonly used in hierarchical clustering computes the Euclidean distance
between rows of the adjacency matrix, namely
\begin{equation}
\label{eqdiss2}
\delta_{ij} = \sqrt{\sum_{k=1}^N (a_{ik} - a_{jk})^2}.
\end{equation} 
For unweighted graphs and ignoring normalization, \eqref{eqdiss1} is equivalent to the square of
\eqref{eqdiss2} since $| \mathcal{N}_i \triangle \mathcal{N}_j | = 
\| \mathbf{a}_i \oplus  \mathbf{a}_j \|_1$, where $\mathbf{a}_i := [a_{i1}, \dots, a_{iN} ]^T$.
A downside to these metrics is their tendency to disproportionately 
assign large dissimilarities to pairs of highly connected nodes 
that do not share single-hop neighbors. 

\subsection{Graph-theoretic distances}
Graph-theoretic distances offer a more compelling option 
for visualization requirements because of their
global nature and ability to capture meaningful relationships between nodes
that do not share immediate neighbors. The \emph{shortest-path} distance is
popular and has been used for many graph embedding approaches because of
its simplicity and the availability of algorithms for its determination. 
It is important to note that it assigns the same distance to node pairs separated by
multiple shortest paths as those with a single shortest path 
yet intuition suggests that node pairs separated by multiple
shortest paths are more strongly connected and therefore more similar.

This shortcoming is alleviated by dissimilarity measures based
on Markov-chain models of random-walks, which have become increasingly important
for tasks such as collaborative recommendation (see e.g., \cite{fouss}). Nodes
are viewed as states and a transition probability, $p_{ij} = a_{ij}/ (\sum_{j=1}^N a_{ij})$,
is assigned to each edge.
The \emph{average commute time} is defined as the quantity 
$n(i,j) = m(j|i) + m(i|j)$, where $m(j|i)$ denotes the \emph{average
first-passage time}, and equals the average number of hops taken 
by a random walker to reach node $j$ given that the starting 
node was $i$. Although $n(i,j)$ is a recursion defined in terms 
of transition probabilities, it admits the following closed form \cite{fouss} 
\begin{equation}
\label{eq21b}
n(i,j) = \Omega \left( \e_i - \e_j \right)^{T} \L^{\dagger}\left( \e_i - \e_j \right) 
\end{equation}
where $ \Omega = \sum_{i,j} a_{ij} $. Moreover, taking the 
spectral decomposition, $\mathbf{L}^{\dagger} = \mathbf{U \Lambda U}^T$, and 
expressing the canonical basis vector,
$\mathbf{e}_i$, as $\mathbf{e}_i = \mathbf{Uz}_i$ with 
$\mathbf{z}_i' := \mathbf{\Lambda}^{1/2} \mathbf{z}_i$, yields
\begin{equation}
\label{eq21c}
[n(i,j)]^{1/2} = \Omega \| \mathbf{z}_i' - \mathbf{z}_j'  \|_2.
\end{equation}
Evidently, $[n(i,j)]^{1/2}$ is an
Euclidean distance in the space spanned by the 
vectors $\{ \mathbf{e}_i \}_{i=1}^N$, each corresponding
to a node in $\mathcal{G}$, and is known as the \emph{Euclidean commute-time
distance (ECTD) } (see \cite{fouss}). For the graph embedding task, setting
$\delta_{ij} = [n(i,j)]^{1/2}$ is convenient because $\mathcal{G}$
is treated as $N$-dimensional Euclidean space with a well-defined criterion
for measuring pairwise dissimilarities.

\section{Numerical Experiments}
\label{experiments}

\begin{center}
\begin{table*}[ht]
{
\hfill{}
\begin{tabular}{ | l || l | c | c |}
  \hline                        
 \textbf{Graph} & \textbf{Type} & \textbf{Number of vertices} & \textbf{ Number of edges} \\ \hline
  Karate club graph & undirected & 34 & 78\\ \hline
  London tube (underground) & undirected & 307 & 353 \\ \hline
  ArXiv general relativity (AGR) network & undirected & 5,242 & 28,980 \\ \hline
  Gnutella-04 & directed & 10,879 & 39,994  \\ \hline  
  Gnutella-24 & directed & 26,518 & 65,369 \\ \hline  
\end{tabular}}
\hfill{}
\caption{Networks used for numerical experiments.}
\label{tb:datasets}
\end{table*}
\end{center}

In this section, the proposed approaches are used to visualize a number of
real-world networks shown in Table \ref{tb:datasets}. Appropriate
centrality measures are selected to highlight particular structural properties
of the networks.

\subsection{Visualizing the London Tube}
\label{ltube} 

The first network under consideration is  
the London tube, an underground train transit
network\footnote{https://wikis.bris.ac.uk/display/ipshe/London+Tube}
whose nodes represent stations whereas the edges
represent the routes connecting them. 
A key objective of the experiment is a graph
embedding that places stations traversed by most routes
closer to the center, thus highlighting their relative significance
in metro transit. 

\begin{figure}[t!]
\centering
\includegraphics[scale=0.5]{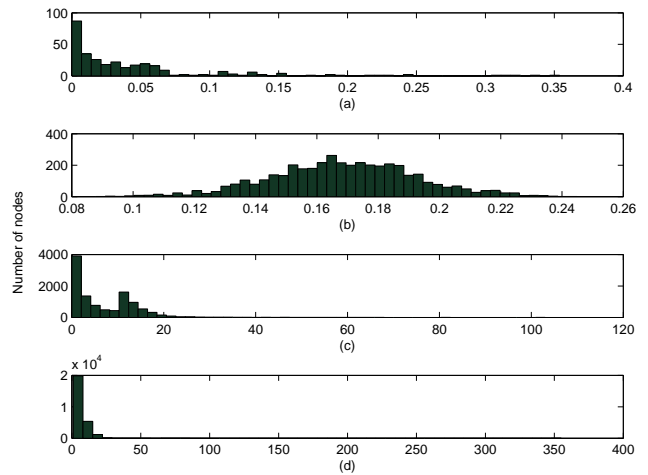}
\caption{$50$-bin centrality histograms for a) London tube network (betweenness centrality), 
 b) AGR network (closeness centrality), c) Gnutella-04 network (degree), and d) Gnutella-24 network (degree). }
\label{fig:histograms}
\end{figure}

Such information is encoded
by the betweenness centralities of the tube stations. It is worth mentioning
that pursuit of this goal deviates from traditional metro transit map requirements
which, albeit drawn using geographically inaccurate schemes, visually emphasize
accessibility between adjacent stations. Figure
\ref{fig:histograms}a) shows the betweenness centrality histogram computed 
for the London tube. 
As expected from measurements performed on real networks, the distribution
follows a power law with a small number
of stations disproportionately accounting for the higher centrality values.
Knowledge of the centrality distribution guides the selection 
of a transformation function, $f(c_i)$, that will result in a well-spaced 
graph embedding. Bearing this in mind, the centrality values were 
transformed as follows:
\begin{equation}
\label{eqq2} f\left( c_i \right) = \frac{\text{diam} \left(
\mathcal{G} \right) }{2} \left( 1 - \frac{c_i - \underset{i \in
\mathcal{V}}{\operatorname{min}} \text{ } c_i}{ \underset{i \in
\mathcal{V}}{\operatorname{max}}\text{ }c_i - \underset{i \in
\mathcal{V}}{\operatorname{max}}\text{ }c_i} \right)
\end{equation}
with $\text{diam}( \mathcal{G})$ denoting the diameter of
$\mathcal{G}$. Node dissimilarities were computed via
ECTD.

\begin{figure*}[t!]
\centering
\begin{tabular}{|c||c|c|c|}
\hline
Algorithm & Betweenness centrality & Closeness centrality & Degree centrality \\
\hline
& & &  \\
CC-MDS
&
\begin{minipage}{120pt}
\centerline{\includegraphics[width=3.75cm]{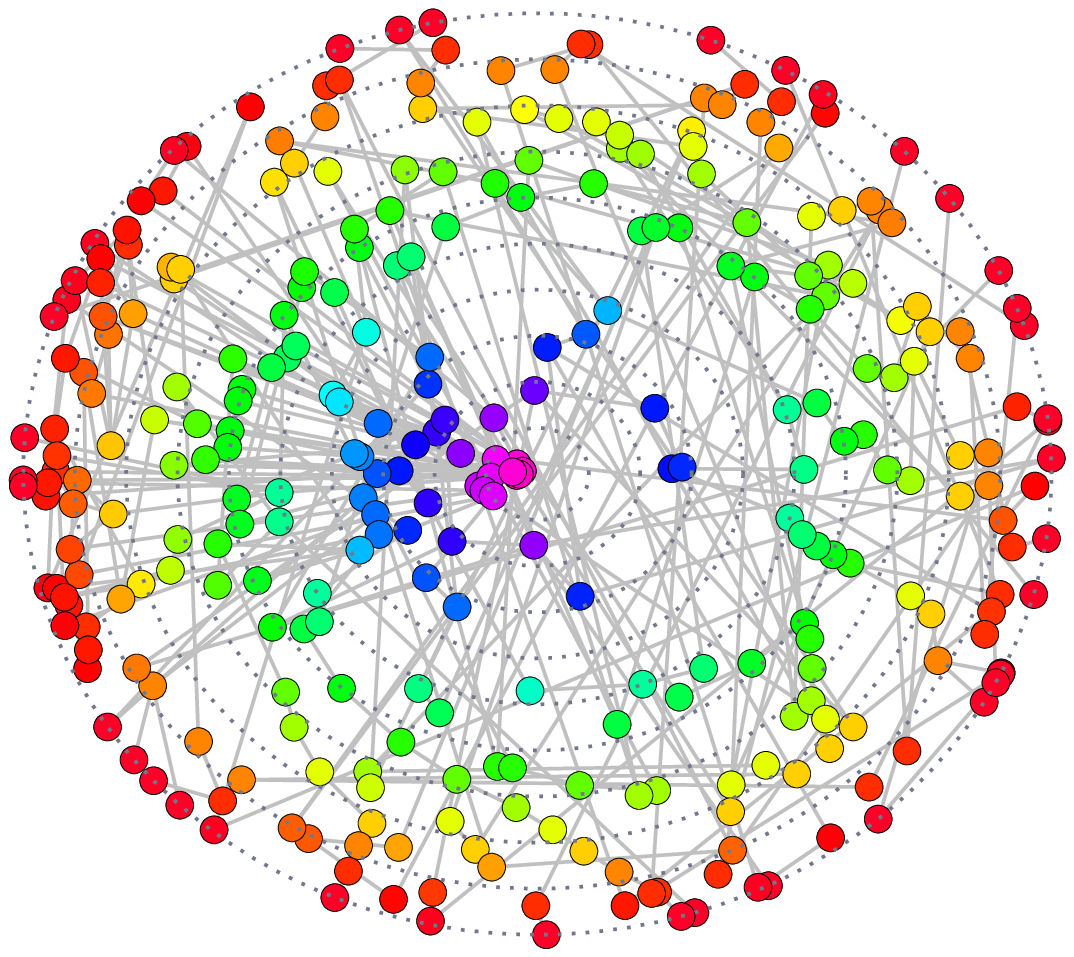}}
\end{minipage}
&
\begin{minipage}{120pt}
\centerline{\includegraphics[width=3.75cm]{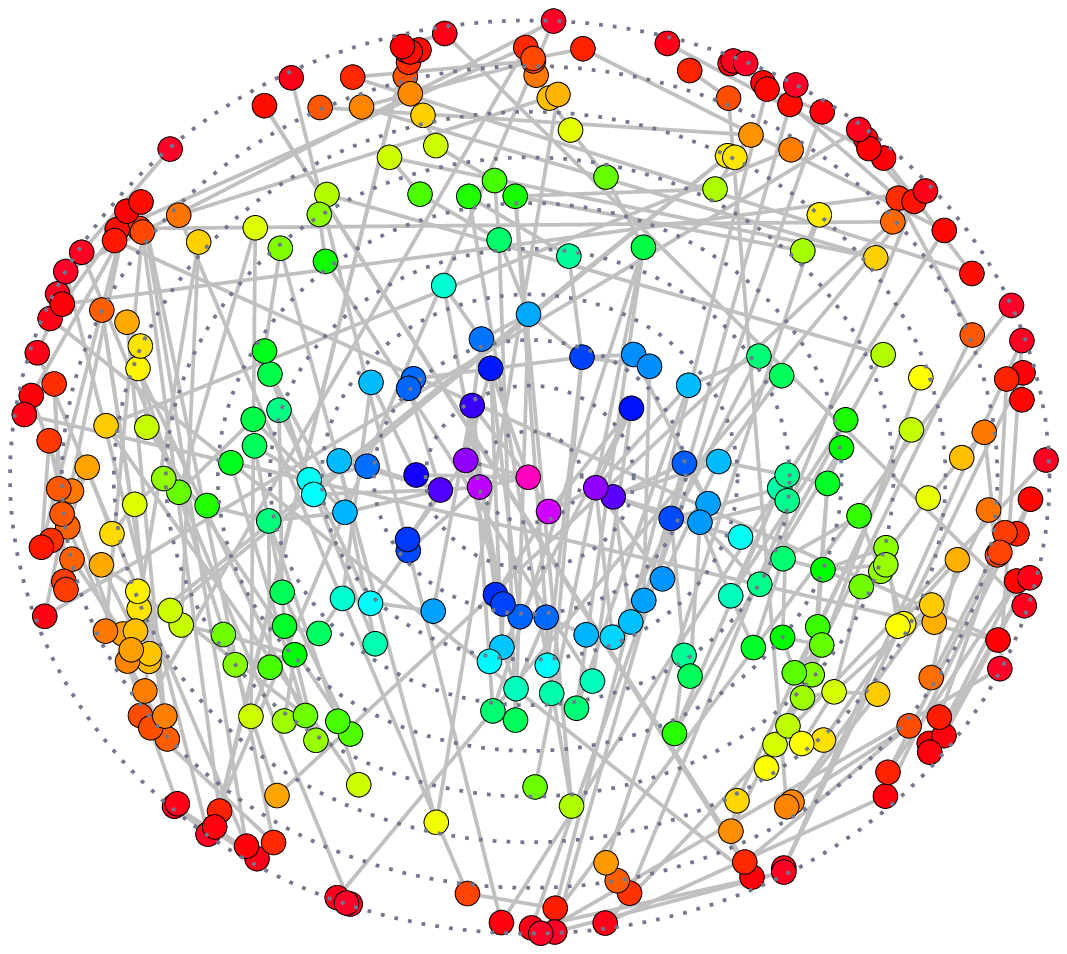}}
\end{minipage}
&
\begin{minipage}{120pt}
\centerline{\includegraphics[width=3.75cm]{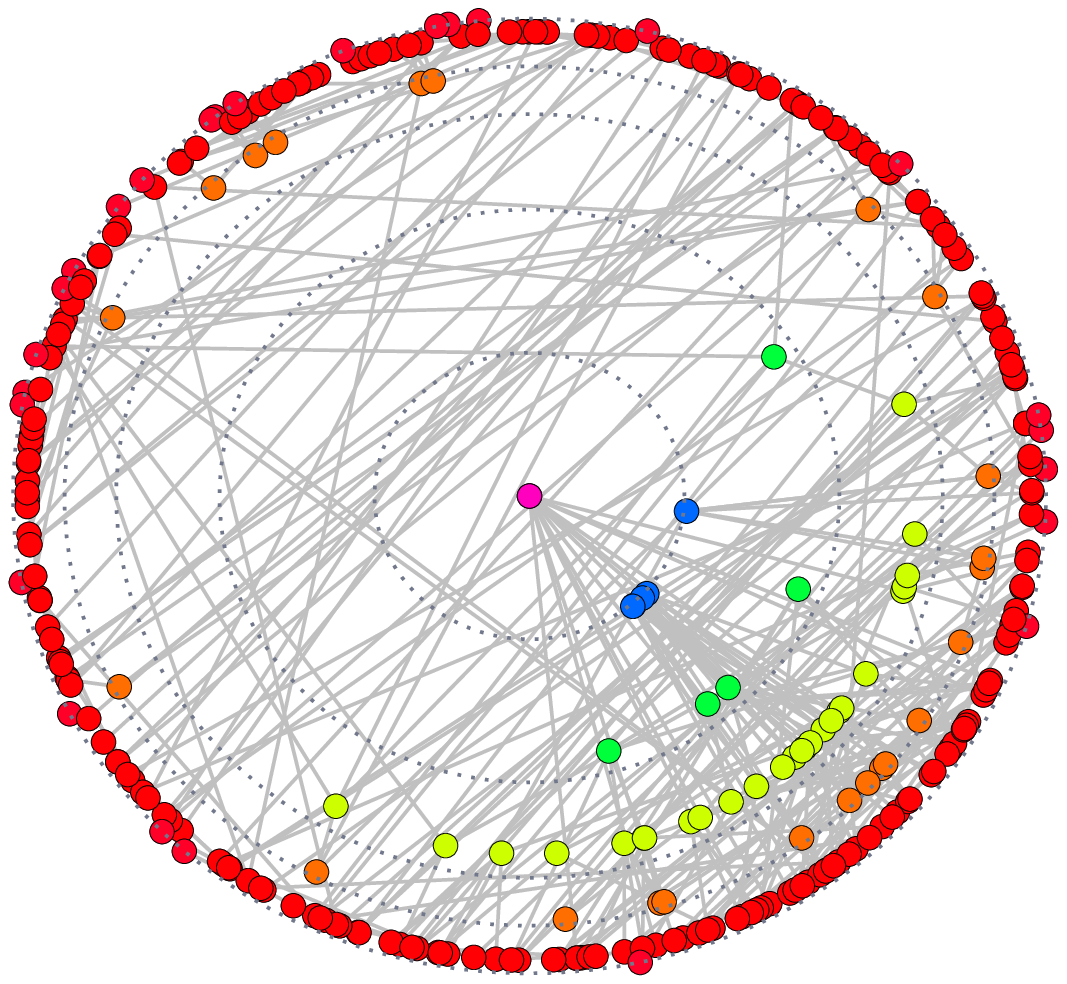}}
\end{minipage}
\\ 
& & &  \\
\hline
& & &  \\
CC-LLE
&
\begin{minipage}{120pt}
  \centering
  \centerline{\includegraphics[width=3.75cm]{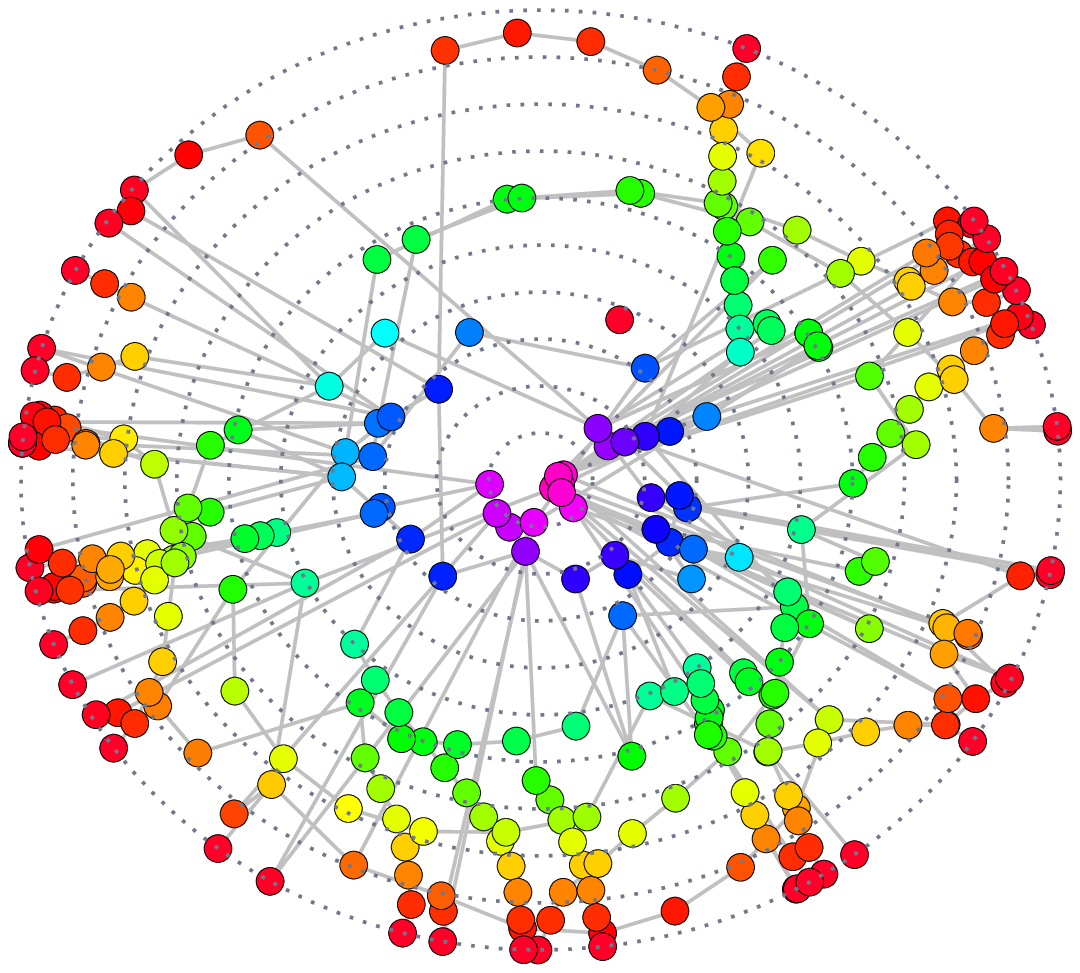}}
\end{minipage}
&
\begin{minipage}{120pt}
  \centering
  \centerline{\includegraphics[width=3.75cm]{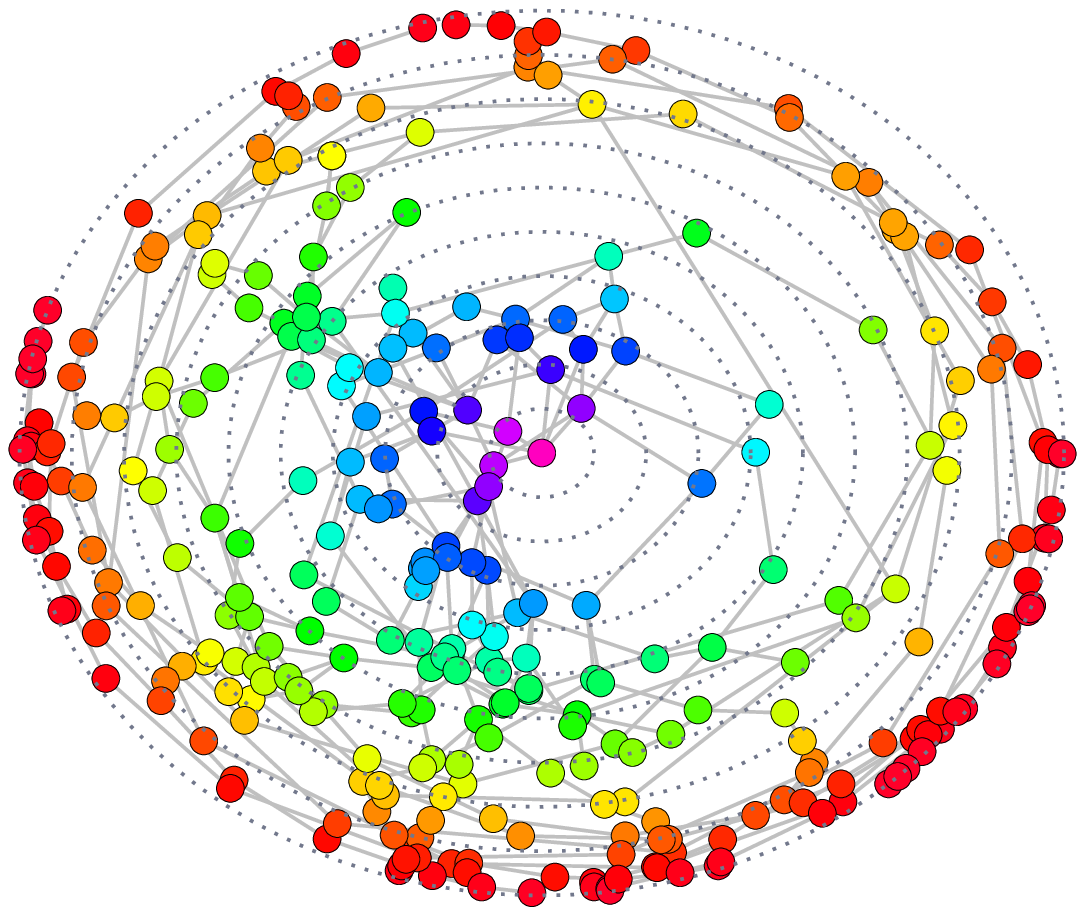}}
\end{minipage}
&
\begin{minipage}{120pt}
  \centering
  \centerline{\includegraphics[width=3.75cm]{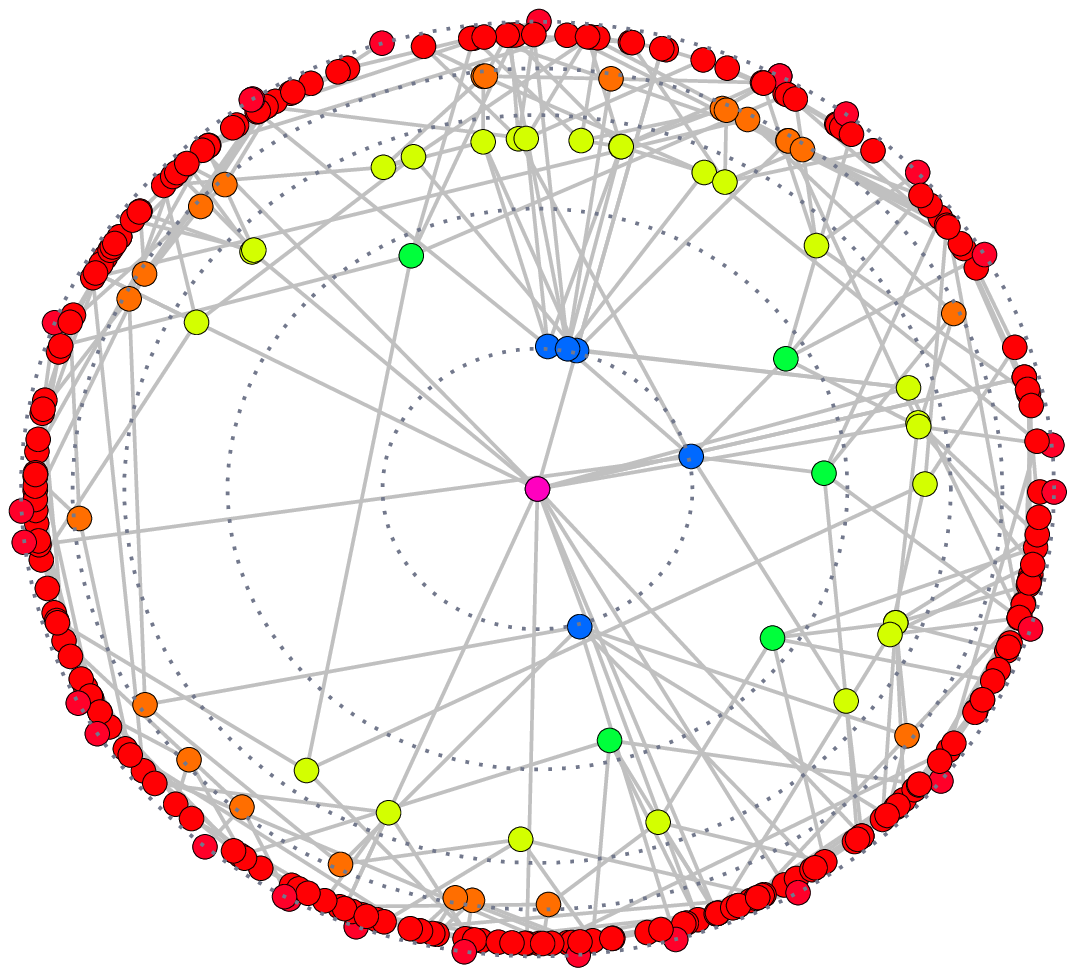}}
\end{minipage}
\\
& & &  \\
 \hline
& & &  \\
MFL
&
\begin{minipage}{120pt}
  \centering
  \centerline{\includegraphics[width=3.75cm]{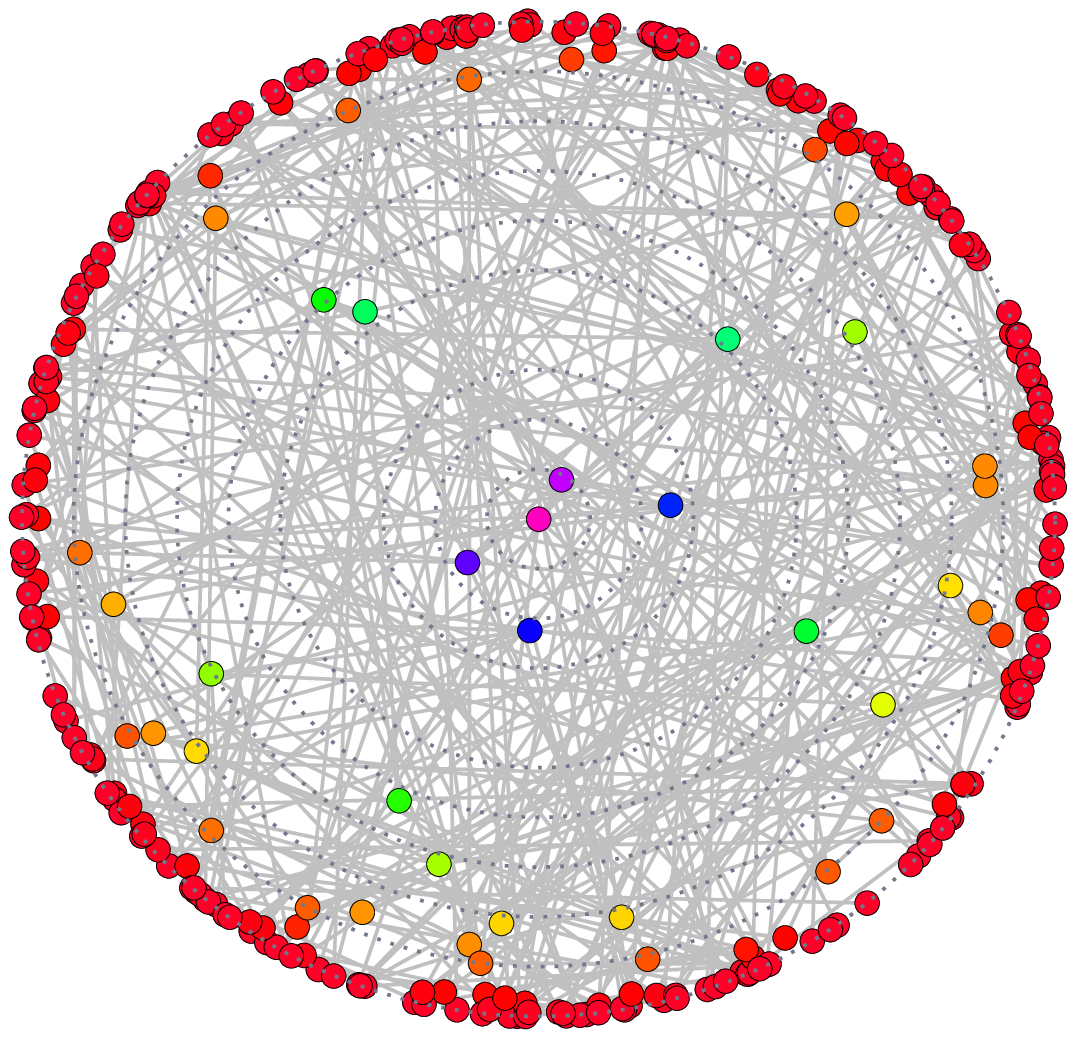}}
\end{minipage}
&
\begin{minipage}{120pt}
  \centering
  \centerline{\includegraphics[width=3.75cm]{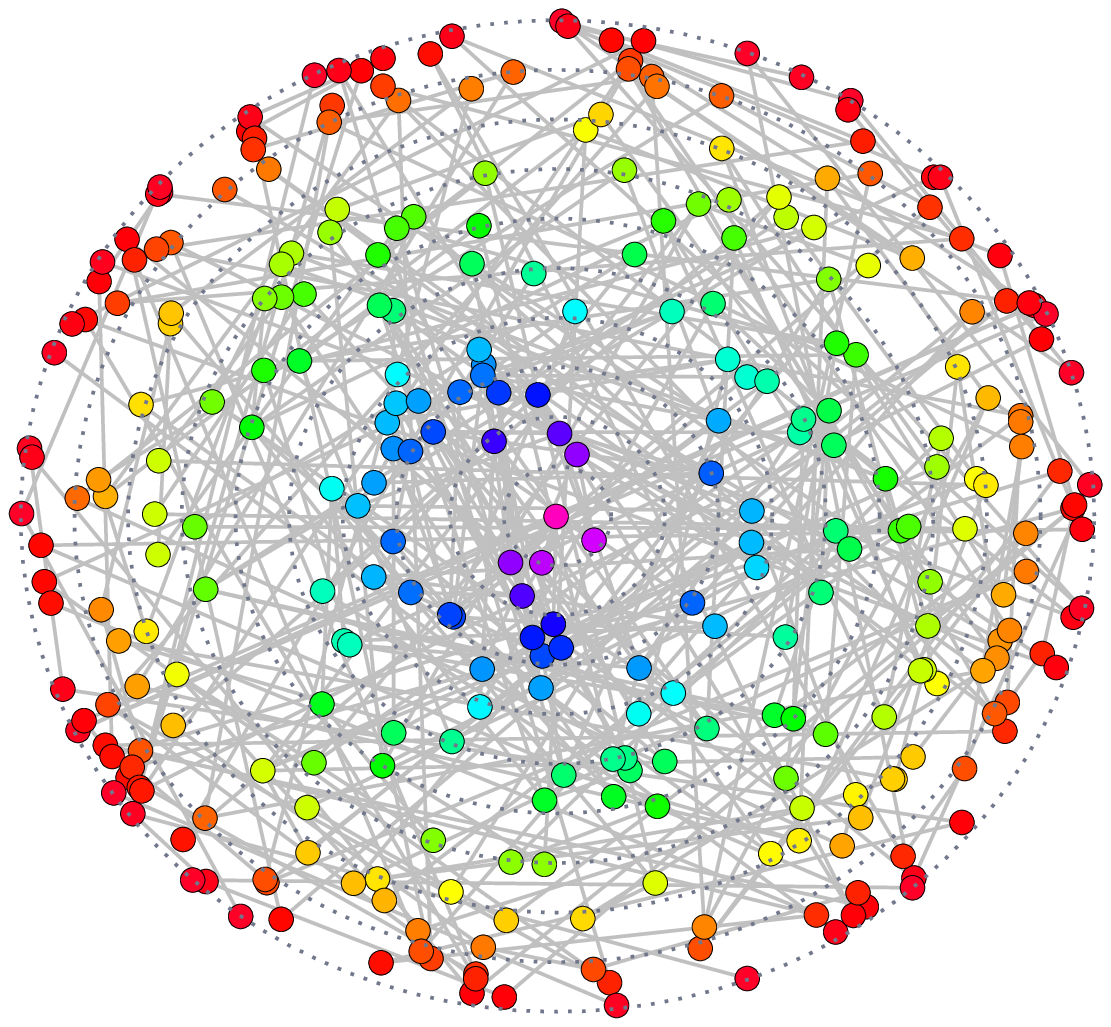}}
\end{minipage}
&
\begin{minipage}{120pt}
  \centerline{\includegraphics[width=3.75cm]{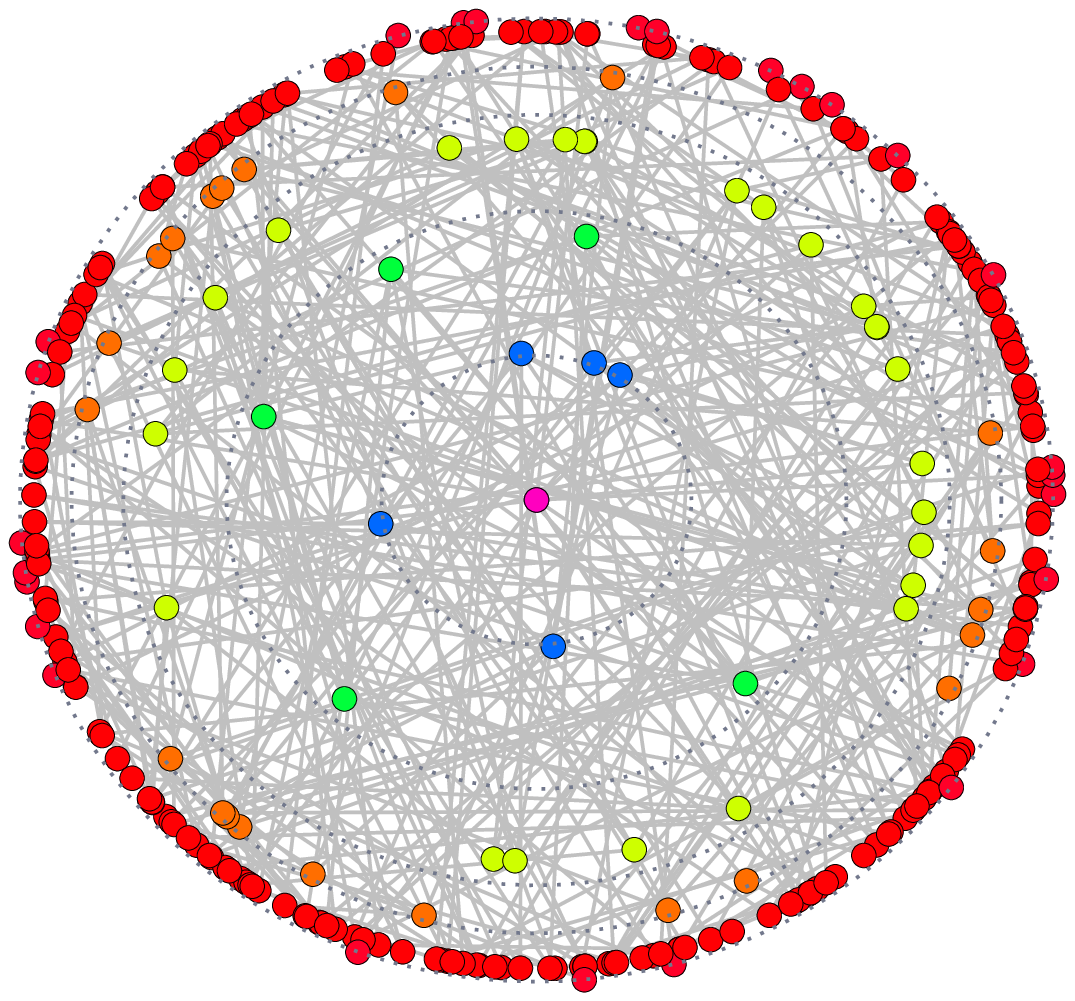}}
\end{minipage}
\\
& & & \\
\hline
& & &  \\
SE
&
\begin{minipage}{120pt}
  \centering
  \centerline{\includegraphics[width=3.75cm]{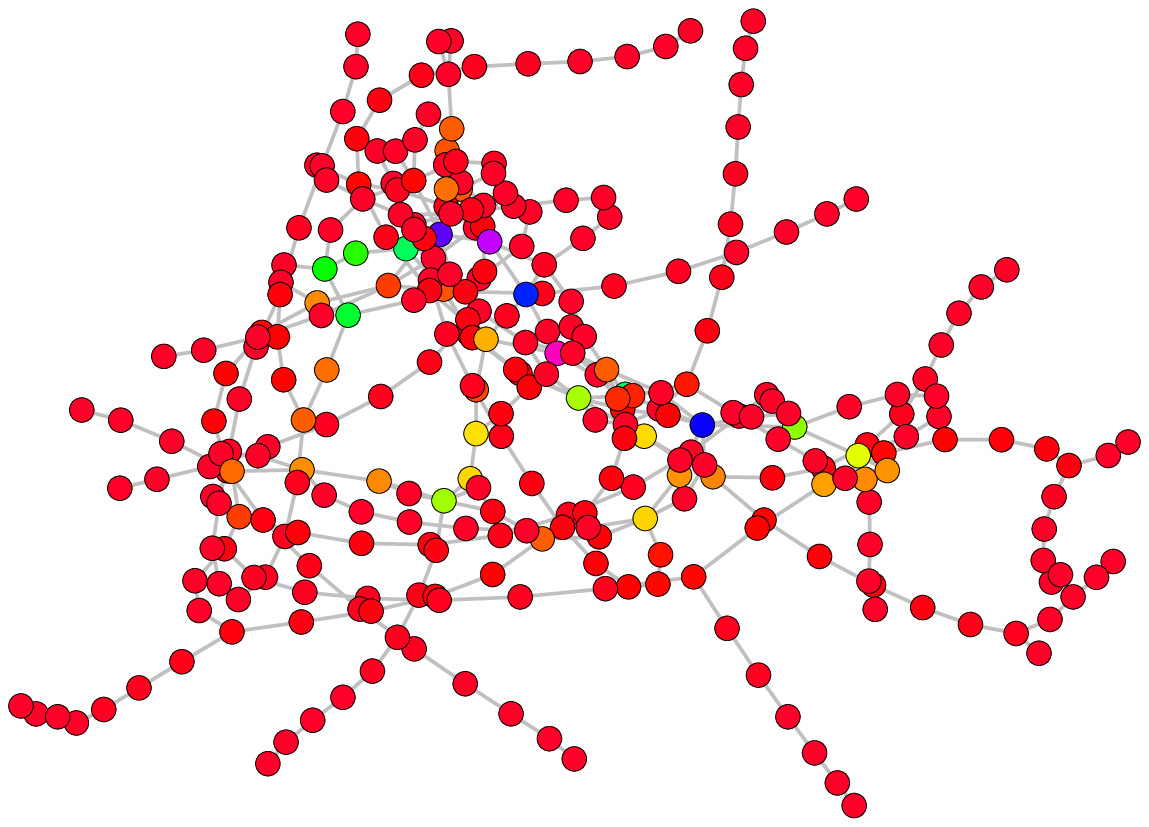}}
\end{minipage}
&
\begin{minipage}{120pt}
  \centering
  \centerline{\includegraphics[width=3.75cm]{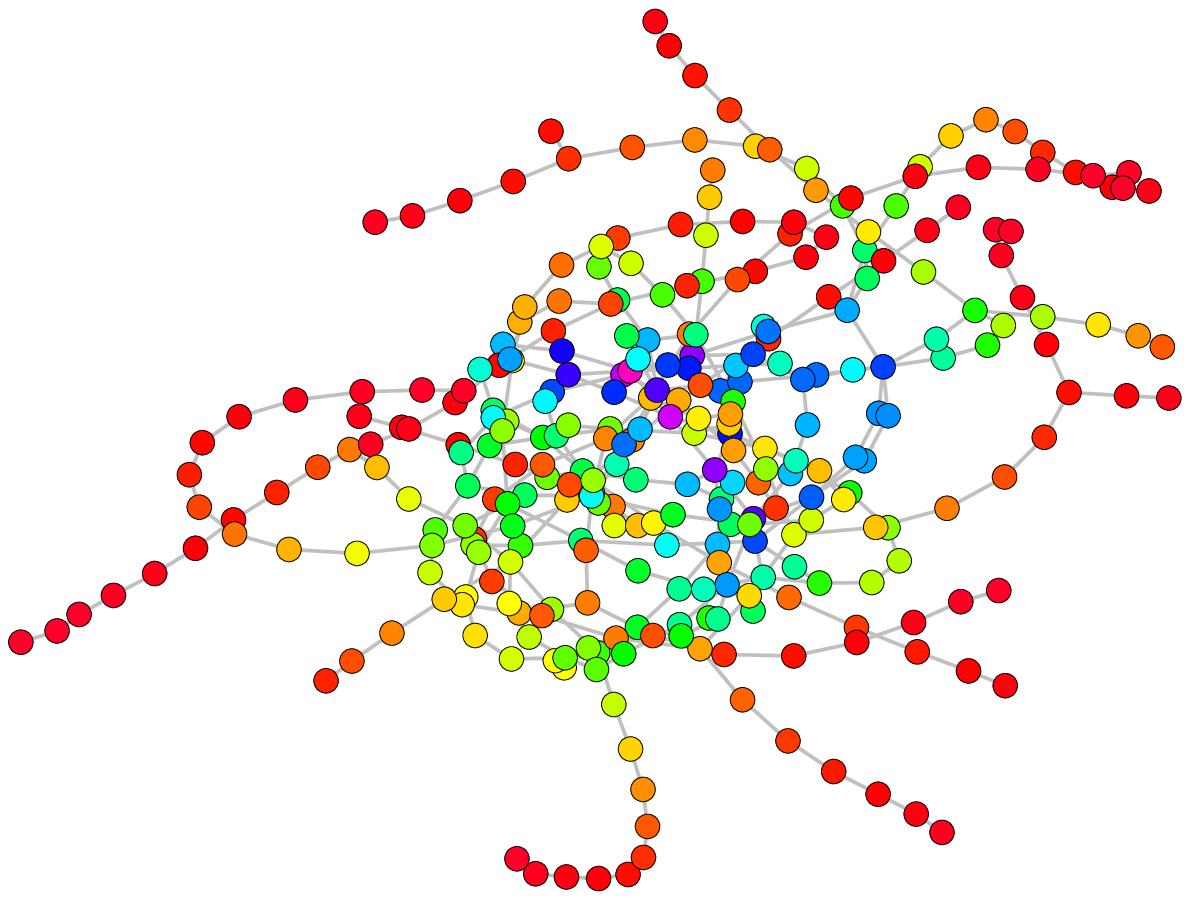}}
\end{minipage}
&
\begin{minipage}{120pt}
  \centerline{\includegraphics[width=3.75cm]{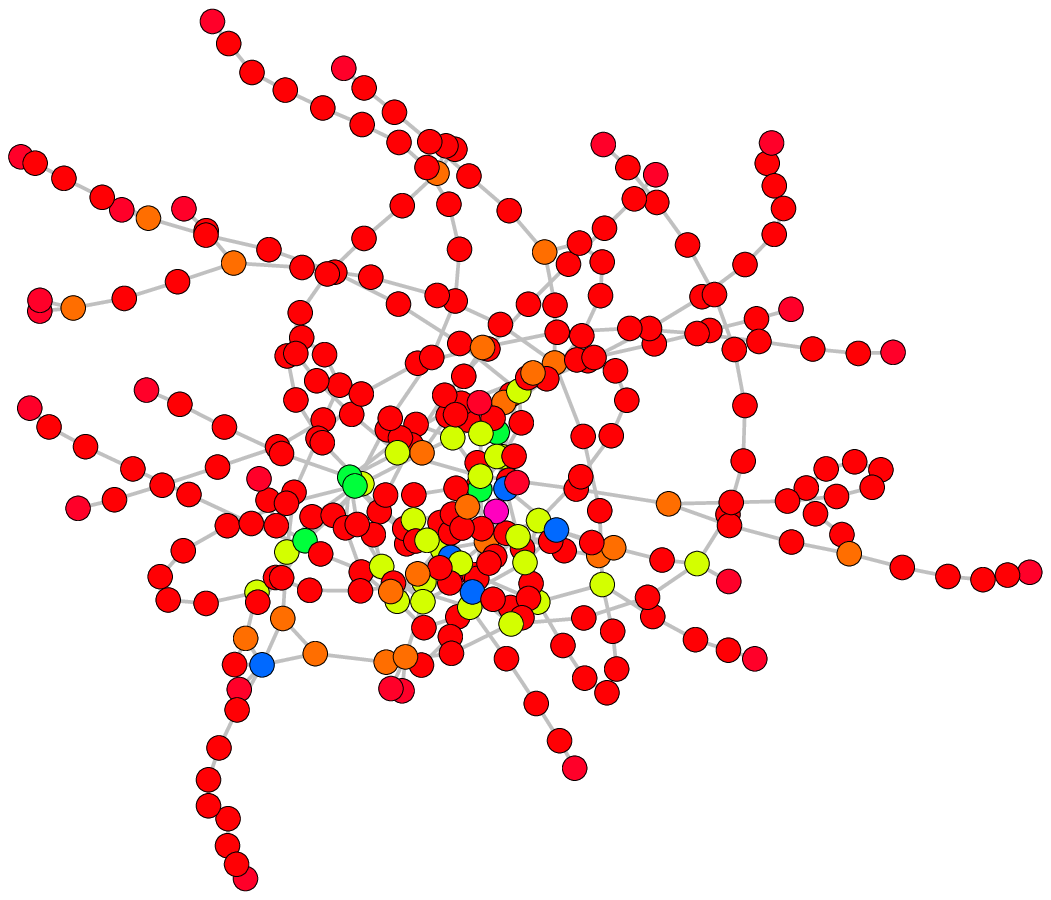}}
\end{minipage}
\\
& & &  \\ 
\hline
\end{tabular}
\caption{Embedding of the London tube graph emphasizing different centrality considerations.}
\label{fig:different_centralities}
\end{figure*}

The visual quality of the embeddings from CC-MDS and CC-LLE 
was evaluated against a recent
centrality-based graph drawing algorithm, a \emph{"More Flexible Radial Layout"} (MFRL) \cite{brandes},
 and the \emph{spring embedding} (SE) \cite{fruch} algorithm.
 The first column in Figure \ref{fig:different_centralities}
depicts the embeddings obtained by running CC-MDS (without a smoothness penalty), 
CC-LLE, MFRL and SE emphasizing betweenness
centrality. Interesting comparisons were drawn by running the four algorithms with 
closeness centrality and node degree input data. The resultant embeddings are depicted
in the last two columns of the figure.

\begin{figure}[t!]
\centering
\includegraphics[scale=0.5]{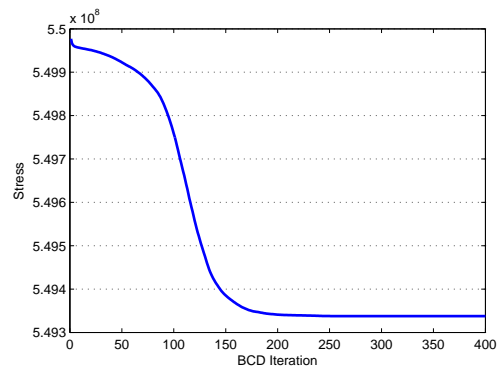}
\caption{MDS stress vs. iterations.}
\label{fig:fig_london_tube_stress_vs_iterations}
\end{figure}

In order to compute the 
subgradients required in CC-MDS, vector $\mathbf{s}$ was set to $(1/\sqrt{2})[1, 1]^T$ 
in \eqref{eq6}. The color grading 
reflects the centrality levels of the 
nodes from highest (violet) to lowest
(red). In general Algorithm~\ref{alg1} 
converged after approximately $150$ BCD
iterations as depicted in Figure \ref{fig:fig_london_tube_stress_vs_iterations}
plotted for the betweenness centrality embedding.
CC-LLE was 
run by setting $n=1$ (single-hop neighbors) and the second row in Figure \ref{fig:different_centralities}
depicts the resulting embeddings after $20$ BCD iterations in the final
step. For the initialization of  the BCD iterations, $\X^0$ was generated
from a Gaussian distribution of unit covariance matrix, i.e., 
$\x_i^0 \sim \mathcal{N}(\mathbf{0}, \mathbf{I})$ for all $i=1,\dots,N$.

Visual inspection of the the approaches
reveals that CC-LLE inherently minimizes the number of edge-crossings
and yields drawings of superior quality for moderately sized networks. 
This behaviour is a direct result of the implicit constraint that
the embedding vector assigned to each node is only influenced by the
immediate neighbors. However, this benefit comes at the cost of
the lack of a convergence guarantee for the BCD iterations. Although
the MFRL algorithm requires pre-selection of the number of iterations,
the criterion for this selection is unclear and experimental results
demonstrate that visual quality is limited by edge-crossings even for
medium-sized (hundreds of nodes) networks.

The last row in Figure \ref{fig:different_centralities} depicts
the spring embedding for the London tube, a force-directed algorithm
that models the network as a mass-spring system and the
embedding task amounts to finding the positions in which the balls
will settle after the system has been perturbed. 
Despite the success of force-directed methods for the
visualization of small networks, it is clear that their ability to 
communicate meaningful hierachical structure rapidly degrades even
for moderately sized networks.

\subsection{Minimizing edge crossings}
\begin{figure}[t!]
\begin{minipage}[b]{0.48\linewidth}
  \centering
  \centerline{\includegraphics[width=3.5cm]{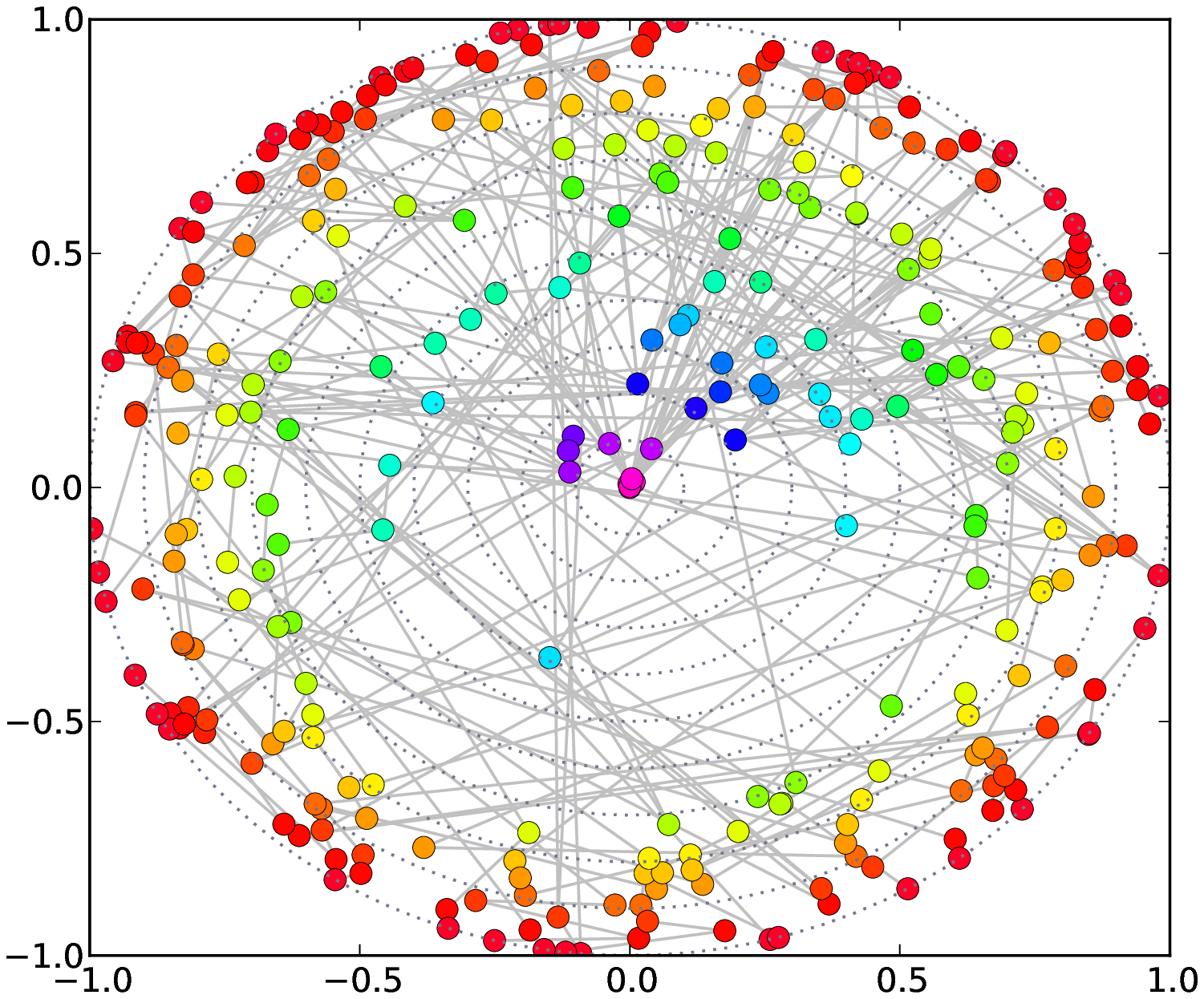}}
  \centerline{ \footnotesize{ (a) $\lambda = 0$}}\medskip
\end{minipage}
\hfill
\begin{minipage}[b]{0.48\linewidth}
  \centering
  \centerline{\includegraphics[width=3.5cm]{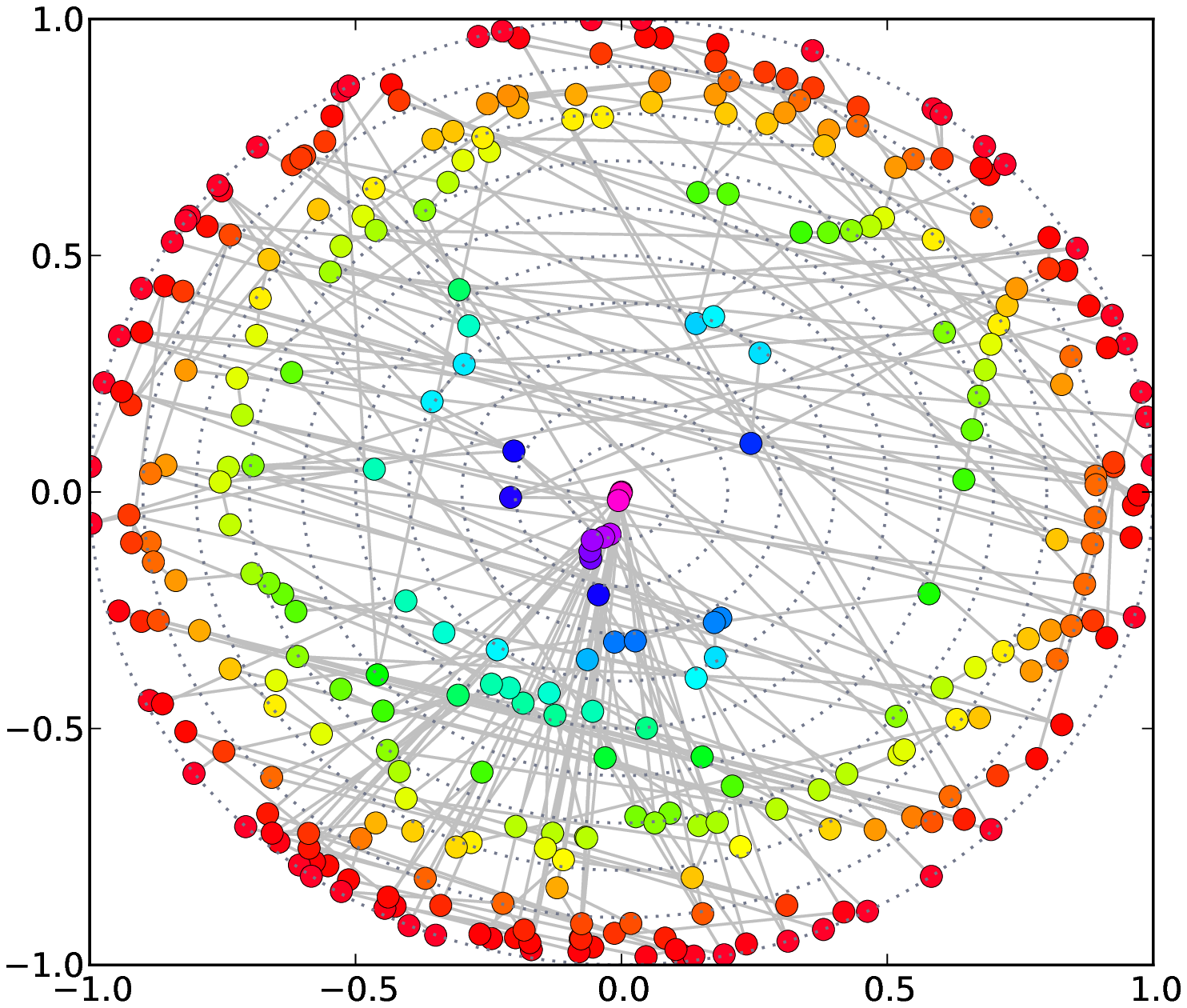}}
  \centerline{ \footnotesize{ (b) $\lambda = 1$}}\medskip
\end{minipage}
\begin{minipage}[b]{0.48\linewidth}
  \centering
  \centerline{\includegraphics[width=3.5cm]{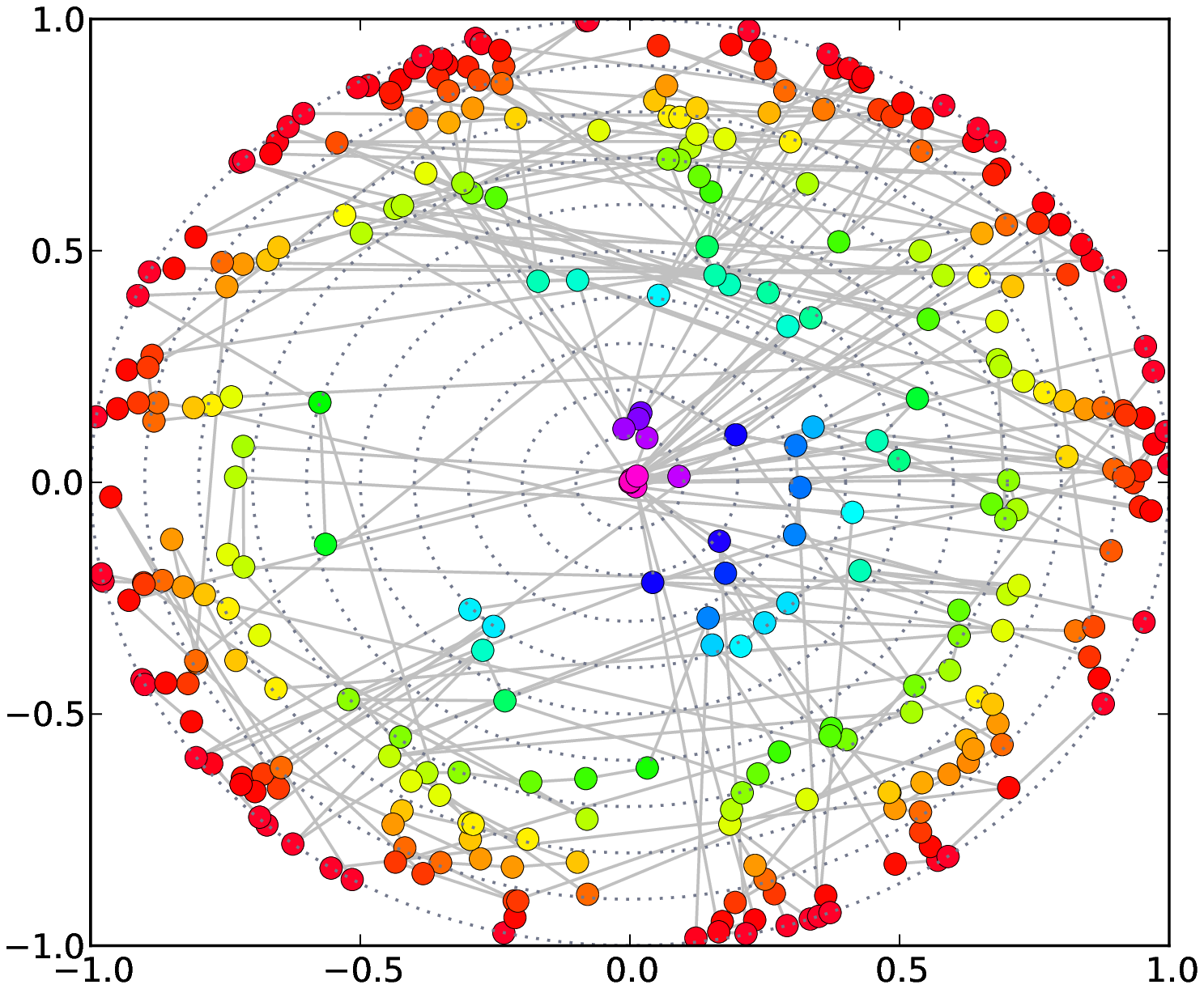}}
  \centerline{\footnotesize{ (c) $\lambda = 100$} }\medskip
\end{minipage}
\hfill
\begin{minipage}[b]{0.48\linewidth}
  \centering
  \centerline{\includegraphics[width=3.5cm]{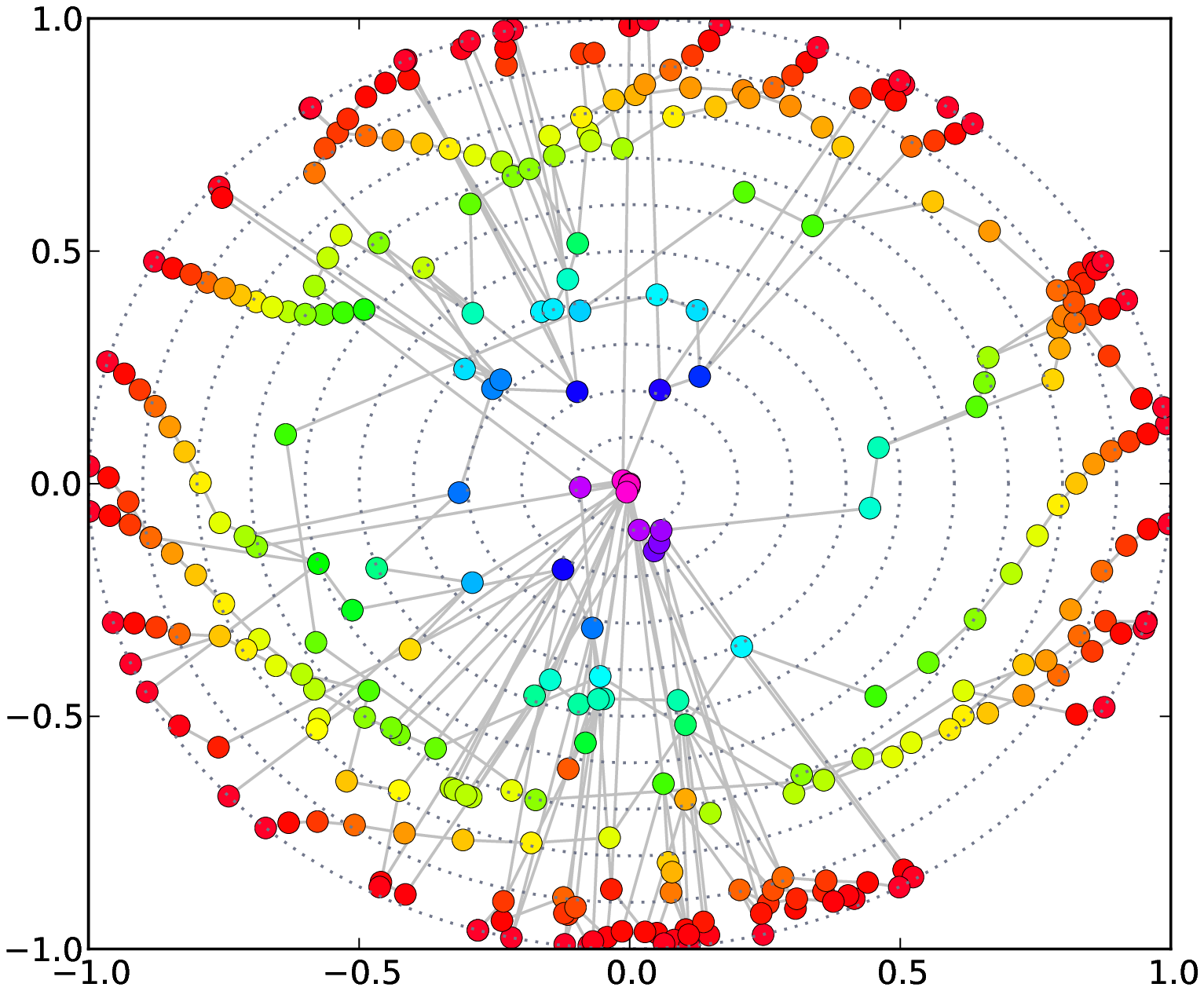}}
  \centerline{\footnotesize{ (d) $\lambda = 10,000$} }\medskip
\end{minipage}
\caption{Effect of a smoothness penalty: varying $\lambda$ for the London tube visualization.}
\label{fig:varying_lambda}
\end{figure}

Turning attention to Algorithm~\ref{alg2}, simulations 
were run for several values of $\lambda$
starting with $\lambda = 0$, and the resultant 
embeddings were plotted as shown in Figure \ref{fig:varying_lambda}.
Increasing $\lambda$ promotes embeddings in
which edge crossings are minimized. This intuitively makes sense
because forcing single-hop neighbors to lie close to each other,
decreases the average edge length, leading to fewer edge crossings.
In addition, increasing $\lambda$ yielded embeddings that were
aesthetically more appealing under fewer iterations.  For instance,
setting $\lambda = 10,000$ required only $30$ iterations for a
visualization that is comparable to running $150$ iterations with
$\lambda = 0$. 

\begin{figure}[!htb]
\centering
\includegraphics[scale=0.45]{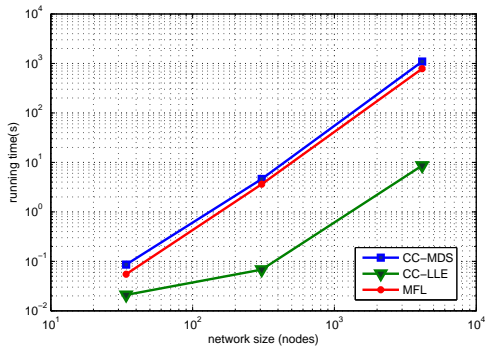}
\caption{A comparison of running times.}
\label{fig:rtimes} 
\end{figure}

\subsection{Visualization of large networks}
\label{arxiv} The primary focus in this subsection is large-scale network 
visualization and test results on three networks of varying size are considered.
First among these is the \emph{ArXiv General Relativity (AGR)} network,
representing scientific
collaborations between authors on papers submitted to the ``General
Relativity and Quantum Cosmology'' category (January
$1993$ to April $2003$)~\cite{leskovec}. The nodes represent 
authors and an edge exists between nodes $i$ and $j$ if authors 
$i$ and $j$ co-authored a paper. Closeness centrality 
was used to visualize the network so that authors whose research is most related to the
majority of the other authors are placed closer to the center. The distribution of 
closeness centrality for the AGR network is shown in 
Figure~\ref{fig:histograms}b). Node dissimilarities
were based on pairwise shortest path distances between
nodes.

\begin{figure*}[t!]
\begin{minipage}[b]{0.48\linewidth}
  \centering
  \centerline{\includegraphics[scale=0.42]{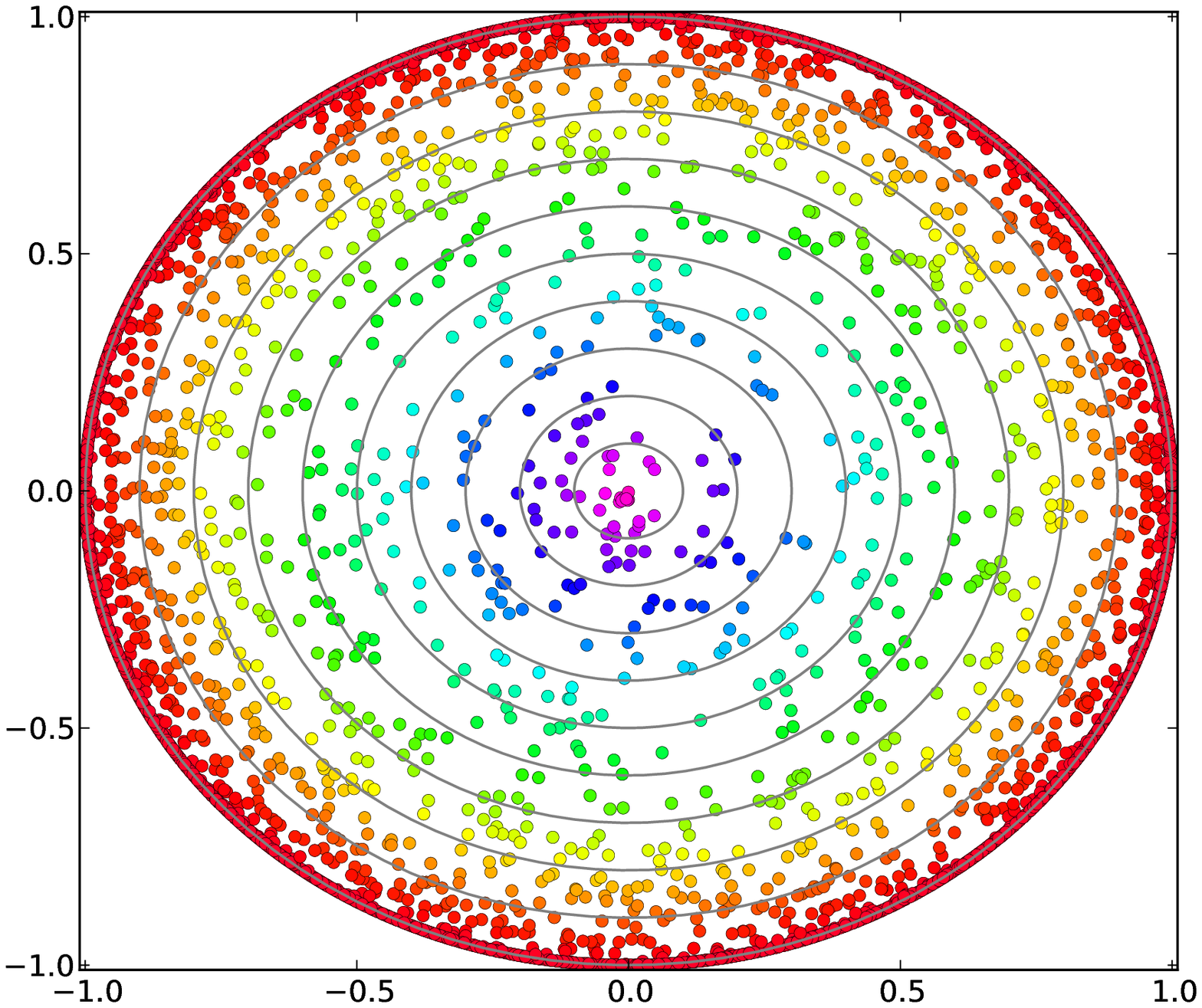}}
  \centerline{ \footnotesize{ (a) }}\medskip
\end{minipage}
\hfill
\begin{minipage}[b]{0.48\linewidth}
  \centering
  \centerline{\includegraphics[scale=0.4]{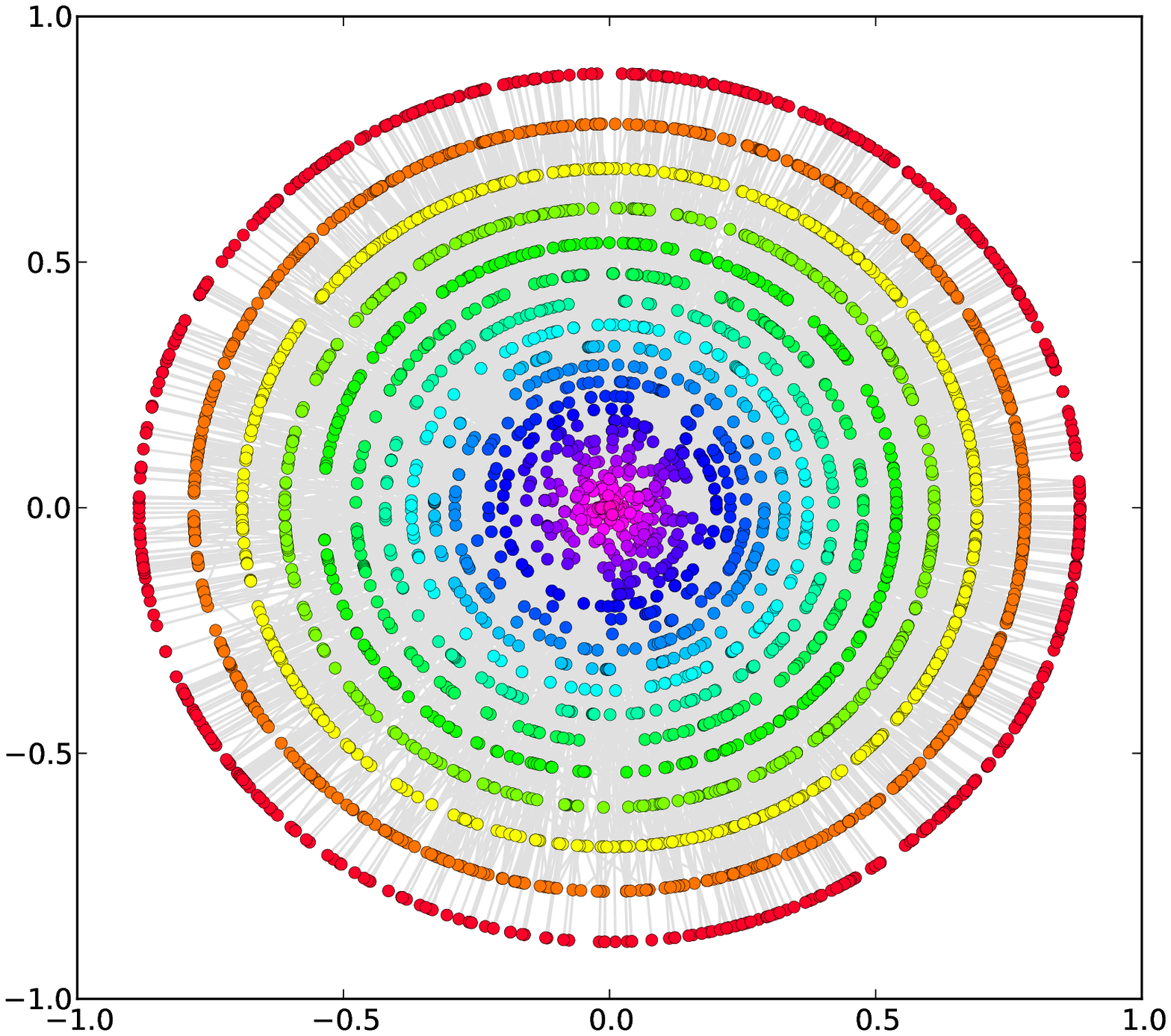}}
  \centerline{ \footnotesize{ (b) }}\medskip
\end{minipage}
\begin{minipage}[b]{0.48\linewidth}
  \centering
  \centerline{\includegraphics[scale=0.4]{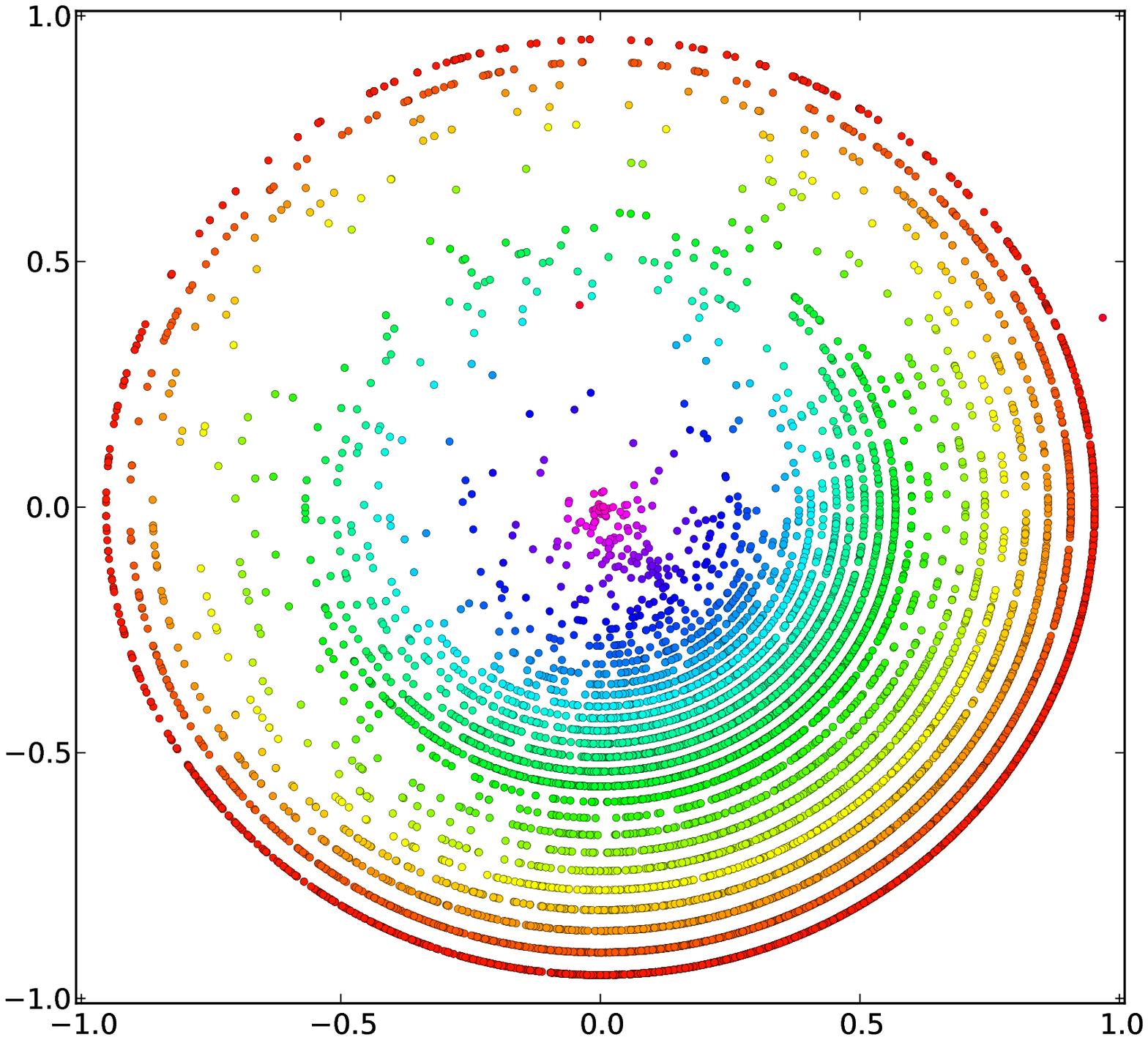}}
  \centerline{\footnotesize{ (c) } }\medskip
\end{minipage}
\hfill
\begin{minipage}[b]{0.48\linewidth}
  \centering
  \centerline{\includegraphics[scale=0.42]{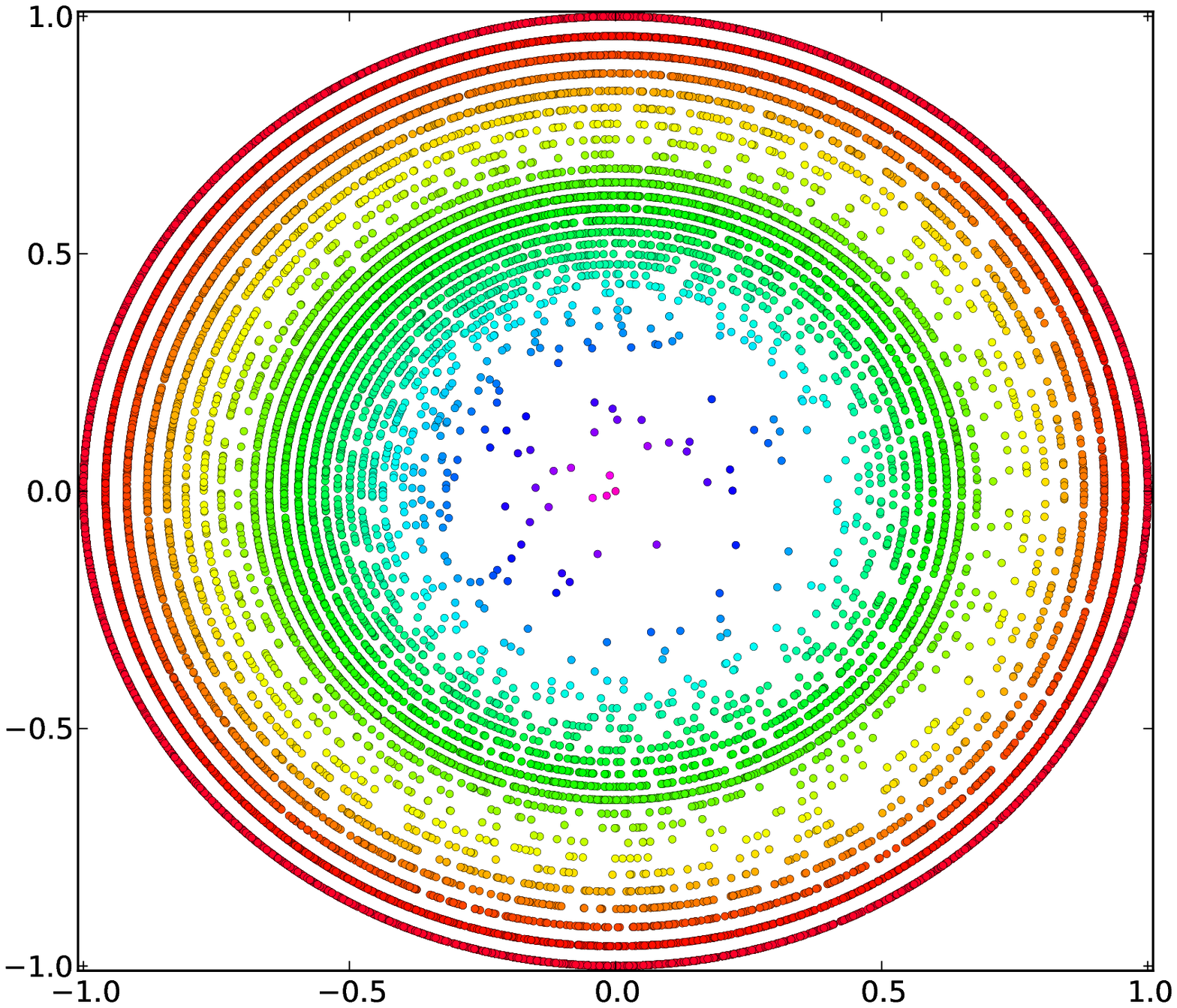}}
  \centerline{\footnotesize{ (d)} }\medskip
\end{minipage}
\caption{Visualization of large networks: a) AGR (CC-MDS), b) AGR (CC-LLE), c) Gnutella-04 (CC-LLE), and d) Gnutella-24 (CC-LLE).}
\label{fig:large_nets}
\end{figure*}

Figure \ref{fig:large_nets}a) shows the embedding 
obtained after $30$ outer BCD iterations of running 
Algorithm~\ref{alg1}. For clarity and emphasis of the node positions,
edges were not included in the visualization. Drawings of graphs as
large as the autonomous systems within the Internet typically thin
out most of the edges. The color coding is adopted to
reflect centrality variations. 

CC-LLE was run for the same data yielding the network
visualization depicted by Figure \ref{fig:large_nets}b). Only single-hop
neighbors were considered ($n=1$) for the neighborhood selection. Although
no claims of optimal placement or convergence are made for Algorithm 
\ref{alg3}, it is clear that the visualization succeeds at conveying
the centrality structure of the collaboration network. 

Next, large-scale degree visualizations of snapshots of the Gnutella peer-to-peer
file-sharing network \cite{leskovec} were generated using CC-LLE. Nodes represent
hosts and edges capture connections between them. Gnutella-$04$ and Gnutella-$24$ represent 
snapshots of  the directed network captured in $2002$ on August $4$  and August $24$ respectively. 
Undirected renditions of the two networks were obtained by symmetrization of their
adjacency matrices (i.e., $\mathbf{A} \gets \mathbf{A} + \mathbf{A}^{\top} $).
The centrality metric of interest was the node degree and dissimilarities for the two visualizations
were computed based on the number of shared neighbors between any pair of hosts. 
Figures \ref{fig:large_nets}c) and \ref{fig:large_nets}d) depict the visualizations obtained via CC-LLE. It is clear that
despite the dramatic growth of the network over a span of $20$ days, most new nodes
had low degree and the number of nodes accounting for the highest degree was disproportinately
small as shown in the histograms in Figures \ref{fig:histograms}c) and \ref{fig:histograms}d).

\subsection{Running times}
Running times for CC-MDS, 
CC-LLE and MFRL were compared for three of the networks
of different dimensions. These include the karate club network ($34$ nodes), a small
social network from a $1970$s study of long-term interactions between 
members of a U.S. university karate club.
In addition, the London tube and the AGR network were considered. Figure  \ref{fig:rtimes}
shows a log-plot of the run-times that led to embeddings
with reasonably comparable visual quality. The main observation from 
the plot is the clear run-time advantage of CC-LLE over the other methods
especially for large networks. In fact, this was the main motivation for using CC-LLE
for visualizing the Gnutella networks yielding embeddings in 
$1,684$~s (Gnutella-$04$) and $5,639$~s (Gnutella-$24$).

\section{Conclusion}
\label{conclusion}
In this paper, two approaches for embedding graphs with 
certain structural constraints were proposed. In the first approach,
an optimization problem was formulated under centrality 
constraints that capture relative levels of importance
between nodes. A block coordinate descent solver with 
successive approximations was developed to deal with the
non-convexity and non-smoothness of the constrained MDS
stress minimization problem. In addition, a smoothness 
penalty term was incorporated to minimize edge crossings in 
the resultant network visualizations. Tests on real-world 
networks were run and the results demonstrated that 
convergence is guaranteed, and large networks can be visualized
relatively fast.

In the second proposed approach, LLE was adapted to the visualization problem
by solving centrality-constrained optimization
problems for both steps of the algorithm. The first step, which 
determines reconstruction weights per node, amounted to a QCQP that
was solved in closed-form. The final step which determines the 
embedding that best preserves the weights turned out to decouple 
across nodes and was solved via BCD iterations with 
closed-form solutions per subproblem. Despite the lack 
of an optimality guarantee for this step, meaningful 
network visualizations were obtained after a few iterations making 
the developed algorithm attractive for large scale graph embeddings.

From an application perspective, the LLE approach
is preferrable over MDS in settings where the network
represents data sampled from an underlying manifold. For instance, consider
a social network whose nodes represent people and 
edges encode self-reported friendship ties between them. 
It is assumed that friends have a number of similar attributes 
e.g., closeness in age, level of education, and income bracket. 
Although some of these attributes may be unknown,
it is reasonable to assume that they lie
near a low-dimensional nonlinear manifold, nicely motivating
an LLE approach to graph embedding.
Some real-world networks are heterogeneous with multi-typed nodes and edges e.g.,
a bibliographic network that captures the relationships
between authors, conferences and papers. The manifold
assumption does not apply here, but dissimilarity measures 
between multi-typed nodes have been developed \cite{shi},
and an MDS approach is well justified.

In this work, it has been assumed that networks are static and 
their topologies are completely known. However, real-networks 
are dynamic and embeddings must be determined from incomplete
data due to sampling and future research directions will incorporate 
these considerations.

\appendices
\section{Proof of Proposition \ref{prop1}}
\label{appendix:prop1}
Substituting $\x_0 = \x^{r-1}$ in the LHS of \eqref{eq10} yields
\begin{eqnarray}
\label{appendixeq1}
\nonumber
\Phi(\x^{r-1}, \x^{r-1}) &= & \psi_1(\x^{r-1}) - \psi_2'(\x^{r-1}, \x^{r-1}) \\
\nonumber
&= & \frac{(N-1)}{2}\| \x^{r-1} \|_2^2 \\
\nonumber
& & - (\x^{r-1})^T(\sum_{j < i}\x_j^r + \sum_{j > i}\x_j^{r-1}) \\
\nonumber
& & - \sum_{j < i} \delta_{ij} \| \x^{r-1} - \x_j^r \|_2 \\
& & - \sum_{j > i} \delta_{ij} \| \x^{r-1} - \x_j^{r-1} \|_2.
\end{eqnarray}
Since
\begin{eqnarray}
\label{appendixeq2}
\nonumber
\Psi (\x^{r-1}) & = & \psi_1(\x^{r-1}) - \psi_2(\x^{r-1}) \\
\nonumber
&= & \frac{(N-1)}{2}\| \x^{r-1} \|_2^2 \\
\nonumber
& & - (\x^{r-1})^T(\sum_{j < i}\x_j^r + \sum_{j > i}\x_j^{r-1}) \\
\nonumber
& & - \sum_{j < i} \delta_{ij} \| \x^{r-1} - \x_j^r \|_2 \\
& & - \sum_{j > i} \delta_{ij} \| \x^{r-1} - \x_j^{r-1} \|_2,
\end{eqnarray}
condition \eqref{eq10} is satisfied by the equivalence of \eqref{appendixeq1}
and \eqref{appendixeq2}. The second condition, \eqref{eq10b}, is automatically satisfied for 
$\x_0 = \x^{r-1}$ by \eqref{eq5}.

\section{Solving the KKT conditions}
\label{appendix:kkt}
This appendix derives \eqref{eqlle10a} - \eqref{eqlle10e} from the KKT conditions 
in \eqref{eqlle11} - \eqref{eqlle13}. From \eqref{eqlle10e}, one obtains

\begin{equation}
\label{appeq1}
2\H_i \w_i^* - 2\h_i + 2\gamma^{*}\H_i \w_i^* + \mu^{*}\ones = \mathbf{0}.
\end{equation}

\noindent
Solving for $\w_i^*$ in terms of $\gamma^*$ and $\mu^*$ from \eqref{appeq1}

\begin{equation}
\label{appeq1b}
\w_i^* = \frac{1}{(1+\gamma^*)}\H_i^{-1}\left(\h_i - \frac{\mu^*}{2}\ones \right).
\end{equation} 

\noindent
Upon applying the result in \eqref{appeq1b} to the primal feasibility 
condition in \eqref{eqlle10b}, it turns out that

\begin{equation}
\label{appeq2}
\ones^T\left[ \frac{\H_i^{-1}}{(1+\mu^*)} \left( \h_i - \frac{\mu^*}{2} \ones \right) \right] = 1.
\end{equation}

\noindent
Thus,
\begin{equation}
\label{appeq2b}
\gamma^{*} = \ones^T \H_i^{-1}\h_i - \frac{\mu^*}{2} \ones^T \H_i^{-1} \ones - 1.
\end{equation}
Assuming the inequality constraint is inactive at optimality, then $\gamma^* = 0$, which
upon solving \eqref{appeq2b} yields

\begin{equation}
\label{appeq3}
\mu^* = \frac{2\left( \ones^T \H_i^{-1} \h_i - 1 \right)}{\ones^T \H_i^{-1}\ones}.
\end{equation}

\noindent
Combining \eqref{appeq3} and \eqref{appeq1b} gives

\begin{equation}
\label{appeq4}
\w_i^* = \H_i^{-1} \left[ \h_i - \frac{\ones^T \H_i^{-1} \h_i - 1}{\ones^T \H_i^{-1}\ones} \ones \right].
\end{equation}

\noindent
If the inequality constraint is active at optimality, then

\begin{equation}
\label{appeq5a}
\w_i^{*T}\H_i \w_i^* = f^2(c_i).
\end{equation}

\noindent
Substituting for $\w_i^*$ from \eqref{appeq1b} results in

\begin{equation}
\label{appeq5b}
\left( 2\h_i - \mu^* \ones \right)^T \H_i^{-1}
\left( 2\h_i - \mu^* \ones \right) 
= 4f^2(c_i)(1+\gamma^*)^2
\end{equation}

\noindent
which upon simplification yields

\begin{eqnarray}
\label{appeq5c}
\nonumber
\h_i^T \H_i^{-1} \h_i - \mu^* \ones^T \H_i^{-1} \h_i + 
\frac{\mu^{*2}}{4} \ones^T \H_i^{-1} \ones \\
= f^2(c_i) (1+\gamma^*)^2.
\end{eqnarray}

\noindent
Substituting for $\gamma^*$ results in the following quadratic 
equation in $\mu^*$

\begin{eqnarray}
\label{appeq5d}
\nonumber
\h_i^T \H_i^{-1} \h_i - \mu^* \ones^T \H_i^{-1} \h_i + \frac{\mu^{*2}}{4} \ones^T \H_i^{-1} \ones \\ 
= f^2(c_i) \left( \ones^T \H_i^{-1}\h_i - \frac{\mu^*}{2} \ones^T \H_i^{-1} \ones \right)^2.
\end{eqnarray}

\noindent
Finally, solving for $\mu^*$ from \eqref{appeq5d} and taking one root yields
\begin{eqnarray}
\label{appeq6}
\nonumber
\mu^* = 2\frac{\ones^T\H_i^{-1}\h_i}{\ones^T\H_i^{-1}\ones}\\
+ 2 \left\lbrace \frac{\h_i^T\H_i^{-1}(\ones\ones^T\H_i^{-1}\h_i - \h_i\ones^T\H_i^{-1}\ones)}{(\ones^T\H_i^{-1}\ones)^2 - (\ones^T\H_i^{-1}\ones)^3 f^2(c_i)} \right \rbrace^{\frac{1}{2}}.
\end{eqnarray}


\vspace*{-5cm}

\end{document}